\documentclass[letterpaper]{article} 
\usepackage{aaai24}  
\usepackage{times}  
\usepackage{helvet}  
\usepackage{courier}  
\usepackage[hyphens]{url}  
\usepackage{graphicx} 
\usepackage{natbib}  
\usepackage{caption} 
\usepackage{algorithm}
\usepackage{algorithmic}

\usepackage{newfloat}
\usepackage{listings}
\DeclareCaptionStyle{ruled}{labelfont=normalfont,labelsep=colon,strut=off} 
\floatstyle{ruled}
\newfloat{listing}{tb}{lst}{}
\floatname{listing}{Listing}
\nocopyright 
\usepackage{booktabs}
\usepackage{subcaption}

\title{Subjectivity in Unsupervised Machine Learning Model Selection}
\author {
    Wanyi Chen\textsuperscript{\rm 1},
    Mary Cummings\textsuperscript{\rm 2}
}
\affiliations {
    \textsuperscript{\rm 1}Duke University\\
    \textsuperscript{\rm 2}George Mason University\\
    wc151@duke.edu, cummings@gmu.edu
}

\begin{document}
\maketitle

\begin{abstract}
Model selection is a necessary step in unsupervised machine learning. Despite numerous criteria and metrics, model selection remains subjective. A high degree of subjectivity may lead to questions about repeatability and reproducibility of various machine learning studies and doubts about the robustness of models deployed in the real world. Yet, the impact of modelers' preferences on model selection outcomes remains largely unexplored. This study uses the Hidden Markov Model as an example to investigate the subjectivity involved in model selection. We asked 33 participants and three Large Language Models (LLMs) to make model selections in three scenarios. Results revealed variability and inconsistencies in both the participants’ and the LLMs' choices, especially when different criteria and metrics disagree. Sources of subjectivity include varying opinions on the importance of different criteria and metrics, differing views on how parsimonious a model should be, and how the size of a dataset should influence model selection. The results underscore the importance of developing a more standardized way to document subjective choices made in model selection processes.
\end{abstract}

\section{Introduction}

In a world of abundant data, unsupervised machine learning (ML) is popular for discovering patterns and structures in data without needing labels. Selecting the best model is a necessary step, as even slightly different models can lead to different interpretations and decisions. For instance, psychologists can use unsupervised ML to uncover patterns in human learning behaviors \cite{visser2002fitting}. Different interpretations of such models would likely lead to different training designs, with some less optimal than others.

There are a few objectives in selecting the best model in unsupervised ML. On the one hand, a model should accurately describe the data. On the other, a parsimonious model is often more desirable because it is more interpretable and less likely to overfit. However, trade-offs exist between accuracy and parsimony. A model with more variables is likely more accurate but less parsimonious \citep{dziak2020sensitivity}. 

Several criteria guide model selection, including information criteria (IC) such as the Akaike Information Criterion (AIC) \citep{akaike1974new} and Bayesian Information Criterion (BIC) \citep{schwarz1978estimating}. Cross-validation metrics such as the most consistent case, best worst case, and best average case may provide additional guidance. 

Despite numerous seemingly objective criteria and metrics, model selection remains subjective. Different ICs place varying emphasis on parsimony versus predictive accuracy \citep{dziak2020sensitivity}. ML researchers and practitioners hold various beliefs about the importance of certain criteria or metrics, so the definition of the ``best" model becomes subjective. While it is well known that biases in datasets cause biases in machine learning models \citep{buolamwini2018gender}, little is known about how the modelers’ preferences influence model selection results. The subjective decisions made in the model selection process can be considered the modeler's ``degree of freedom."

The modeler's degree of freedom yields insights into the repeatability and reproducibility of results, essential in scientific research \citep{plesser2018reproducibility}. If results are not repeatable or reproducible, the research community cannot critically assess the correctness of scientific claims made and conclusions drawn from the results \citep{plesser2018reproducibility}. Consequently, we cannot have confidence in the robustness of various real-world ML models based on these research. 

This study investigates the subjectivity involved in model selection using the Hidden Markov Model (HMM) as an example of unsupervised ML. In the following sections, we first review the background on HMM, information criterion, model selection, and researcher degrees of freedom. Then, we describe an experiment where 33 participants and three Large Language Models (LLMs) were asked to select models in three scenarios.

\section{Background and related work}
An HMM is a stochastic model with two layers: a lower layer of observations and a higher layer of hidden states that is not directly observable. Mathematically, an HMM consists of a transition matrix that represents the transition probabilities among $n$ different hidden states, an emission matrix that represents the emission probabilities from $n$ hidden states to $m$ different observations, and a vector that represents the initial probability distributions among the $n$ hidden states. An HMM's number of free parameters, $n_p$, is calculated by counting the freely-estimated entries in the transition and emission matrices. HMMs with more hidden states are more complex and have larger $n_p$ values.

The Baum-Welch algorithm is typically used to train HMMs \cite{rabiner1989tutorial}. The number of hidden states is unknown and must be chosen before training. Given the same dataset, an algorithm can train different models with different numbers of hidden states.

\subsection{Information criterion}

\begin{table*}[t]
    \centering
    \begin{tabular}{ lccc }
    \toprule
    IC & Penalty term & Emphasis \\
    \midrule
    AIC & $2n_p$ & Predictive accuracy \\
    AIC3 & $3n_p$ & Predictive accuracy \\
    AICc & $(2Tn_p)/(T - n_p - 1)$ & Predictive accuracy \\
    AICu & $(2Tn_p)/(T - n_p - 1) + T \log (T/(T - n_p - 1))$ & Predictive accuracy \\
    CAIC & $n_p (\log T + 1)$ & Parsimony \\
    BIC & $n_p \log T$ & Parsimony \\
    ABIC & $(n_p \log (T+2))/24$ & Depends on $T$ \\
    \bottomrule
\end{tabular}
    \caption{Summary of ICs}
    \label{summary_ic}
\end{table*}

Information criteria, calculated for each candidate model, are often used in model selection. In theory, the candidate model with the minimum IC value should be selected. Different ICs have different theoretical and practical motivations, but they share a common form: 
\[IC = -2(\log L) + \texttt{penalty term}\]
where $\log L$ stands for the log likelihood of the dataset given the model. It captures how well the model fits the data. The first term $-2(\log L)$ characterizes the model's accuracy: lower values indicate that the model has a higher likelihood. In HMM's case, a model's likelihood can be calculated by the forward algorithm \cite{rabiner1989tutorial}. 

The model is likely to become more accurate with more parameters, so to keep the model parsimonious, the ICs add a penalty term based on $n_p$. The value of the penalty term increases as $n_p$ increases. Some ICs' penalty terms also involve $T$, the dataset size. ICs with smaller penalty terms emphasize predictive accuracy, and ICs with larger penalty terms emphasize parsimony. Table 1 summarizes seven common ICs' penalty terms and emphases. Appendix A includes a more detailed review of the theoretical and practical motivation of these ICs.

Among these seven ICs, AIC, AIC3, AICc, and AICu have lighter penalty weights, making them more likely to choose more complex models and risk overfitting. CAIC and BIC have heavier penalty weights, making them more likely to choose simpler models and risk underfitting. ABIC’s penalty weight is lighter than that of BIC’s and may be lighter or heavier than that of AIC’s depending on $T$ \cite{dziak2020sensitivity}. ABIC’s penalty weight becomes heavier with larger $T$. Because of the different penalty terms, these information criteria often suggest different best models when given the same set of candidate models. 

\subsection{Model selection}
Recognizing that different ICs often lead to different models, some researchers have compared the performance of different ICs, with mixed results. Costa et al. \shortcite{costa2010model} performed a Monte Carlo experiment with HMMs and found that the dataset size, the conditional state-dependent probabilities, and the latent transition matrix are the main factors influencing the IC test results. Bacci et al. \shortcite{bacci2014comparison} conducted another Monte Carlo simulation study with HMMs. They also found that the capability of the different ICs in detecting the true number of hidden states varies case by case. Particularly, lower values of persistence probabilities in the same state, and/or a greater uncertainty in the allocation of the observations to the latent states, may lead to worse performance of some ICs \cite{bacci2014comparison}.

These Monte Carlo experiments assume some ground truth exists. However, in many real-world applications, it is impossible to know the true latent transition matrix or many other salient factors. Therefore, it still remains uncertain whether choosing some ICs over others will yield better model selection results for a particular real-world scenario.

Other researchers propose abandoning the use of ICs altogether. Celeux et al. \shortcite{celeux2008selecting} proposed to use cross-validation
to instead assess the number of hidden states in HMMs, i.e., partitioning the dataset into different folds and iteratively assigning different folds as training and testing sets. The model that maximizes the cross-validated likelihood on testing sets is selected. This method selects models with the best predictive performances \cite{celeux2008selecting}. However, this method is often computationally intensive \cite{pohle2017selecting}, and evidence shows that it tends not to outperform BIC \cite{costa2010model}.

Some researchers have developed pragmatic guides for practitioners engaged in model selection. Dziak et al. \shortcite{dziak2020sensitivity} advised practitioners to choose AIC if underfitting is a concern and choose the more parsimonious BIC if overfitting is more of a concern. Pohle et al. \shortcite{pohle2017selecting} advised that in addition to using the ICs, practitioners should also use their expert knowledge to narrow the range of candidate models and use model validation methods to ensure models accurately represent the data. These pragmatic guides attempt to provide selection structure, but little is known about human preferences in this process.

Beyond HMMs, model section occurs in many other fields and contexts. For example, statisticians are interested in analyzing data clusters by means of mixture distributions, and the ICs can help them select the best mixture models \cite{bozdogan1994mixture}. The ICs are also used for determining the number of significant components in Principle Component Analyses, which has a wide range of applications \cite{bai2018consistency}. For the purposes of this study, we use HMM as an example of unsupervised ML to investigate how human preferences influence model selection.

\subsection{Researcher degrees of freedom}
Researchers' subjective choices in data collection, processing, and analysis cause studies to have different results and sometimes support opposite hypotheses \cite{simmons2011false,huntington2021influence}. Simmons et al. \shortcite{simmons2011false} first coined the term ``researcher degrees of freedom" when they discovered that a seemingly small change in researchers' subjective choices could significantly increase the false-positive rate in psychology studies. Similarly, Huntington-Klein et al. \shortcite{huntington2021influence} found that unreported researcher variation could decrease the validity of study results. 

ML researchers recognize that biases in datasets cause biases in resulting models \cite{buolamwini2018gender}, yet practitioner-induced biases are less discussed. More recently, Cummings et al. \shortcite{cummings2021subjectivity} demonstrated that subjective judgments made in the development and interpretation of ML models could lead to inconsistent results and errors of commission and omission. Still, none have explored the impact of researcher degrees of freedom in unsupervised learning model selection, which is the aim of this study.

\section{Method}
We created three model selection scenarios by training HMMs using publicly available replay files from Star-Craft II \cite{blizzard}, a popular online multiplayer Real-Time Strategy game. We focused on Zerg (a role in the game) and grouped its more than 100 different actions into 20 categories. The sequences of player actions, called replays, are our observation sequences. 

We created three datasets corresponding to three model selection scenarios: a small dataset (scenario 1) comprising 30 replay files, a mid dataset (scenario 2) comprising 490 replay files, and a large dataset (scenario 3) comprising 2980 replay files. All files were randomly sampled for all datasets. For each dataset, we trained 19 candidate HMMs with the number of hidden states ranging from 2 to 20. Then, we calculated the IC values for each candidate model (Appendix B). Table 2 summarizes the best model chosen by each IC in each scenario. The models are denoted by their number of hidden states. For example, model 9 refers to the candidate model with 9 hidden states.

\begin{table}
  \centering
  \begin{tabular}{lccc}
    \toprule
    IC & Scenario 1 & Scenario 2 & Scenario 3\\
    \midrule
    AIC & model 9 & model 19 & \underline{model 20}\\
    AIC3 & \emph{model 6} & model 17 & \underline{model 20}\\
    AICc & model 9 & model 17 & \underline{model 20}\\
    AICu & \emph{model 6} & model 13 & \underline{model 20}\\
    CAIC & \bf{model 2} & \bf{model 4} & \emph{model 13}\\
    BIC & \bf{model 2} & \emph{\underline{model 7}} & \bf{model 14}\\
    ABIC & \underline{model 4} & \emph{\underline{model 7}} & \bf{model 14}\\
    \bottomrule
  \end{tabular}
  \caption{Best model selected by each IC in each scenario. Bold = best average model, italics = most consistent model, underline = best worst-case model.}
  \label{best_models_ic}
\end{table}

Next, for the candidate models chosen by at least one IC, we ran additional cross-validation tests. For example, in scenario 1, candidate models 2, 4, 6, and 9 appeared at least once in the second column of Table 2, so we further investigated these models. We evaluated their negative loglikelihood terms $-2(\log L)$ using 5-fold cross-validation. For scenarios 2 and 3, the datasets are larger, so we ran 10-fold cross-validation instead of 5-fold. We report the full results in Appendix C. 

Table 2 also reflects that the ``best average" metric selects the model with the lowest $-2(\log L)$ term average across the 5 or 10 folds. Lower $-2(\log L)$ term values suggest that the model has higher likelihood. We also calculated the $-2(\log L)$ term's standard deviation (SD) across the folds. The ``most consistent" metric selects the model with the smallest SD. Additionally, for each model, the largest $-2(\log L)$ term value among the folds is considered the ``worst case." The ``best worst case" metric selects the model with the smallest worst-case value. 

Given these models, we designed a self-paced survey to measure modelers' preferences in the three model selection scenarios (Appendix D). First, participants completed an informed consent form and a demographics survey (Appendix E). The demographics survey also included a seven-question risk propensity scale survey that measures an individual's general risk-taking tendencies \cite{meertens2008measuring}. We hypothesized that one's risk propensity might correlate to model selection choices. Then, participants were given background information about HMMs and the ICs. 

Next, participants completed an untimed model selection survey, which gave the participants the three model selection scenarios in randomized order. Participants did not know that the models were trained on StarCraft II data. For each scenario, the survey presents the best models chosen by various ICs, the cross-validation results, and the summaries. The participants also have optional access to the tables in Appendix B by clicking a ``click to expand" option. The participants are then asked to select only one ``best" model based on available information, to indicate which factors influenced their decisions the most, and to briefly explain their reasoning (see Appendix F for the full survey). 

Most participants completed the survey in 20 to 40 minutes. The participants were recruited as volunteers but were each given a \$20 Amazon gift card as an appreciation. This study was approved by the Institutional Review Board.

In addition to the human participants, we also incorporated three LLMs into our study: ChatGPT 3.5, ChatGPT 4.0, and Google Bard. Given that LLMs are trained on a vast corpus of human texts, encompassing numerous research papers on model selection, we hypothesized that the LLMs might emulate the most typical human responses to model selection tasks. Given that model selection is an onerous task, we also wanted to explore whether the LLMs could be helpful tools for assisting in this task. For each model selection scenario, the LLMs were presented with identical information as our human participants, including the dataset size, the best models chosen by various ICs, and a summary of the cross-validation results (Appendix I). The next section presents the results.

\section{Results and discussion}
We recruited a total of 33 participants, all with some ML experience. Ages ranged from 20-45 yrs, with an average of 27 yrs (SD = 5 yrs). The group included 8 women and 25 men. 27 were undergraduate or graduate students majoring in computer science or related fields, and 6 were professionals in technical roles. Appendix G provides more details on the demographics. 

The participants' risk propensity scores ranged from 2.14 to 6.17, with a mean of 3.77 (SD = 1.04). Higher scores indicate more risk-taking. Typical risk propensity scale scores range from 2.00 to 7.00, with a mean of 4.63 (SD = 1.23) \cite{meertens2008measuring}. Our mean scores indicate that our group of participants is considerably more risk-averse than the original group, although we cannot establish statistical significance without access to the original data.

\subsection{Participant model selection preferences}
\begin{figure*}
    \centering
    \includegraphics[scale=0.55]{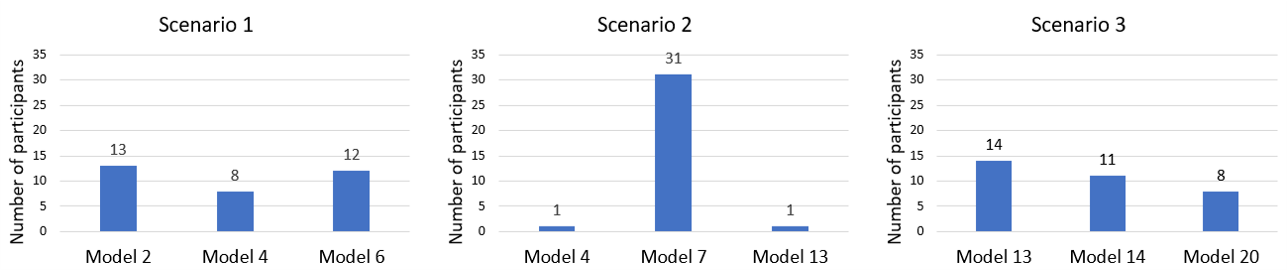}
    \caption{Model selection results}
    \label{fig:result1}
\end{figure*}

In two out of three scenarios, participants’ model selection choices varied. Figure \ref{fig:result1} presents participants’ choices in each scenario. When two of the three cross-validation metrics (best average, most consistent, best worst-case) point to the same model, most participants chose that model. For example, in scenario 2, model 7 is both the most consistent model and the model with the best worst-case. 31 out of 33 participants chose model 7 as the best model. A few participants based their choices on evidence that more than one cross-validation metric agreed on model 7. ``7 seemed like the obvious choice because it was both the most consistent and the best worst case," wrote one participant. Another participant also noted, ``My choice was based primarily on consistency results and best worst case."

However, in the other two scenarios, all three cross-validation metrics point to different models, resulting in significantly more disagreements among participants. Some participants had almost exact opposite reasons for selecting models. For instance, in scenario 1, the simplest HMM possible, a model with two hidden states, is one of the candidate models. Thirteen participants chose this model, with one explaining, ``When it comes to model selection, I often prefer parsimony over accuracy.” Similarly, another participant simply wrote ``the fewest number of states" when asked to explain his or her reasoning. 

On the contrary, some participants refrained from choosing such extreme values. One participant wrote, ``I just don't like 2 because it is the smallest possible number. Explaining everything with just two variables sounds dangerous." Similarly, another participant noted, ``Given 19 observation categories, only 2 latent states (as suggested by BIC) might be too reductionist." A third participant stated, ``I usually do not select models at the boundary of the range, e.g., I don't choose 2 in range 2-20." These different attitudes towards extreme simplicity contributed to disagreements about the best model in this scenario.

Participants also held divergent views on how dataset size should influence model selection. The survey did not explicitly ask how the participants reasoned about dataset sizes, and only six participants (18.2\%) mentioned dataset size as a factor they considered. One participant believed that the ``most consistent" metric was more important for larger dataset sizes, stating, ``Given the large size of the dataset, results on consistency are likely to hold and be a good reflection of the strength of the model." However, another participant held the opposite view, noting, ``If the data size is small, I may prefer a more consistent model." 

There was also disagreement about the importance of accuracy for models trained on small datasets. One person thought accurate predictions were difficult with a small sample size, stating, ``The sample size is really small compared with the previous two datasets. I think it's difficult to get an accurate prediction." Conversely, another participant prioritized accuracy over parsimony for the same scenario, explaining, ``There was not much data, so I selected accuracy over parsimony for information criterion."

Information criteria is another factor that affected participants' selections. Different participants placed importance on different ICs. Most participants indicated that more than one IC was important. Table 3 summarizes their responses. The second column is the number of participants who considered the corresponding IC listed in the first column. Across all three scenarios, BIC is considered most frequently. Other ICs have varying degrees of popularity and are not consistent across scenarios. Among the seven ICs, BIC \cite{schwarz1978estimating} is also the most cited, reaching over 55,000 citations to date. It is possible that the participants considered BIC more frequently because it is more well-known.

Additionally, ten participants (30.3\%) commented that they would like to know more context since it might influence their model selection preferences. Our survey did not give the participants context about the dataset, how it was processed, and the purpose of the model. One participant commented, ``The metrics above are useful in a general sense, but domain knowledge of the problem and how and where the models will be used or deployed also play an important role in model selection."

\begin{table*}[t]
    \centering
    \subfloat[Scenario 1 (small dataset)]{
    \begin{tabular}{lc}
        \toprule
       IC considered &  Count \\
       \midrule
       BIC & 25 \\
       CAIC & 19 \\
       AIC3 & 17 \\
       AICu & 15 \\
       AIC  & 13 \\
       ABIC & 11 \\
       AICc & 6 \\
      \bottomrule
    \end{tabular}}
    \quad
    \subfloat[Scenario 2 (mid dataset)]{
    \begin{tabular}{lc}
        \toprule
       IC considered &  Count \\
       \midrule
       BIC & 30 \\
       ABIC & 26 \\
       CAIC & 14 \\
       AIC  & 10 \\
       AIC3 & 7 \\
       AICc & 6 \\
       AICu & 6 \\
       \bottomrule
    \end{tabular}}
    \quad
    \subfloat[Scenario 3 (large dataset)]{
    \begin{tabular}{lc}
        \toprule
       IC considered &  Count \\
       \midrule
       BIC & 25 \\
       AIC  & 19 \\
       CAIC & 18 \\
       ABIC & 18 \\
       AICu & 15 \\
       AICc & 13 \\
       AIC3 & 12 \\
       \bottomrule
    \end{tabular}}
    \caption{IC considered for each scenario}
\end{table*}

\begin{figure}
    \centering
    \includegraphics[scale=0.77]{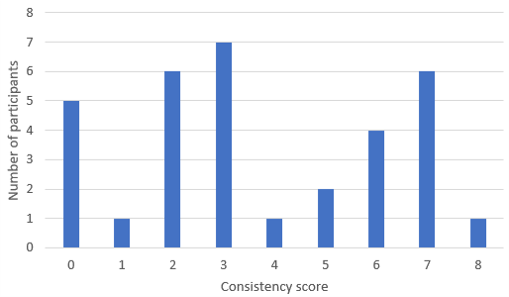}
    \caption{Individual consistency scores (lower scores are more consistent)}
    \label{fig:consistency distribution}
\end{figure}

\begin{table*}[t]
    \centering
    \subfloat[Example survey response of one participant]{
    \begin{tabular}{lccc}
        \toprule
         & Scenario 1 & Scenario 2 & Scenario 3 \\
        \midrule
       a. One or more information criteria  & 4 & 4 & 5\\
       b. Best average performance & 4 & 5 & 4\\
       c. Most consistent performance & 4 & 4 & 4\\
       d. Best worst-case performance & 2 & 2 & 2\\
       \bottomrule
    \end{tabular}}
    \vspace{0.3cm}
    \subfloat[Example pairwise inconsistencies of one participant]{
    \begin{tabular}{lcccc}
        \toprule
         Pairwise comparison & Scenario 1 & Scenario 2 & Scenario 3 & Pairwise inconsistency score\\
        \midrule
       $a$-$b$  & $a=b$ & $a<b$ & $a>b$ & 2\\
       $a$-$c$ & $a=c$ & $a=c$ & $a>c$ & 1\\
       $a$-$d$ & $a>d$ & $a>d$ & $a>d$ & 0\\
       $b$-$c$ & $b=c$ & $b>c$ & $b=c$ & 1\\
       $b$-$d$ & $b>d$ & $b>d$ & $b>d$ & 0\\
       $c$-$d$ & $c>d$ & $c>d$ & $c>d$ & 0\\
       \bottomrule
    \end{tabular}}
    \label{tab:eg_consistency}
    \caption{Example consistency score calculation}
\end{table*}

\subsection{Participant inconsistencies}
We analyzed participants' choice-consistency across scenarios in terms of whether their criteria preferences changed between scenarios. The survey asked each participant to rate the importance of four factors (first column of Table 4(a)) in each scenario. Each factor was rated on a Likert scale of 1 to 5, with 1 being the least important and 5 being the most important. Table 4(a) shows one participant's responses as an example. 

Then, we compared participants' ratings across scenarios. For instance, did participants always think that ``$a.$ one or more information criteria" is more important than ``$b.$ best average performance in cross-validation"? If they rated choice $a$ the same as choice $b$ in one scenario, we denote that as $a = b$. Similarly, $a>b$ means that a participant rated choice $a$ higher than choice $b$, and $a<b$ means that the participant rated choice $b$ higher than choice $a$. If the comparison symbol is consistent across all three scenarios, we denote a pairwise inconsistency score of 0 (the $a$-$d$ row in Table 4(b) offers an example). Similarly, a pairwise inconsistency score of 1 means the comparison symbol in one scenario differs from the other two (the $a$-$c$ row offers an example). A pairwise inconsistency score of 2 means the comparison symbol differs across all three scenarios (the $a$-$b$ row offers an example). 

We added all six pairwise inconsistency scores to obtain an overall consistency score for each participant. In the Table 4 example, the participant's consistency score is 4. Higher scores indicate more inconsistency. The overall mean score is 3.70 (SD = 2.53). Figure \ref{fig:consistency distribution} shows the consistency score distribution. We used k-means clustering to define three clusters. We labeled the six participants (18.2\%) whose scores ranged from 0 to 1 as ``consistent," the fourteen participants (42.4\%) whose scores ranged from 2 to 4 as ``slightly inconsistent," and the thirteen participants (39.4\%) whose scores ranged from 5 to 8 as ``inconsistent."

By this definition, nearly 40\% of participants were inconsistent in their model selection preferences. The cause of the inconsistencies is not clear. Only two of thirteen inconsistent participants explained that they favored different ICs and cross-validation metrics because of different dataset sizes. The others did not explicitly explain why their preferences varied across scenarios, which is in line with behavioral economics research, which has shown that individuals can exhibit inconsistent preferences when presented with the same choice in different forms \cite{kahneman2013prospect}. While no two scenarios are identical in our survey, participants faced similar choices across scenarios, relying on information presented to them in similar forms. The high degree of inconsistency observed in our study suggests that predicting an individual's model selection preferences in a new scenario is challenging, unless many criteria and metrics agree and the choice is relatively straightforward.

\subsection{Participant correlations}
We investigated whether consistency scores correlated with gender, risk attitude, and years of ML experience. For risk attitude, we labeled the six participants (18.2\%) whose risk propensity scale scores were one standard deviation below the mean as ``risk avoiders," the five participants (15.2\%) whose scores were one standard deviation above the mean as ``risk takers," and the remaining as ``neutral." For years of ML experience, the survey asked whether participants had less than one year, one to three years, or more than three years of experience. Since only four participants indicated ``less than one year," we merged the first two experience levels. We clustered the 23 participants (69.7\%) with $\leq 3$ years of experience into one group and the 10 participants (30.3\%) with $>3$ years of experience into another group. 

Then, the Fisher's exact test determined whether nonrandom associations exist. Fisher's exact test is an alternative to the $\chi^2$-test and is recommended when the sample size is small \cite{freeman1951note}. We found no correlations between consistency scores and gender (p = 0.88), risk attitudes (p = 0.78), and years of ML experience (p = 0.34) at the significance level of 0.05.

We also ran Fisher's exact test to investigate whether one's gender, risk attitude, years of ML experience, and individual consistency correlate with one's preferences in model selection. Scenario 2 is omitted for this analysis because 31 out of 33 participants chose the same model. We report the results and p-values in Appendix H. We found no correlations at the significance level of 0.05.

\subsection{LLM model selection preferences}

\begin{table*}[t]
    \centering
    \subfloat[Scenario 1 (small dataset)]{
    \begin{tabular}{lccc}
        \toprule
       LLM & $1^{st}$ try & $2^{nd}$ try & $3^{rd}$ try \\
       \midrule
       ChatGPT 3.5 & model 2 & model 6 & model 6 \\
       ChatGPT 4.0 & model 2 & model 6 & model 4 \\
       Google Bard & model 2 & model 2 & model 6 \\
      \bottomrule
    \end{tabular}}
    \quad
    \subfloat[Scenario 2 (mid dataset)]{
    \begin{tabular}{lccc}
        \toprule
       LLM & $1^{st}$ try & $2^{nd}$ try & $3^{rd}$ try \\
       \midrule
       ChatGPT 3.5 & model 7 & model 4 & model 7 \\
       ChatGPT 4.0 & model 7 & model 7 & model 7 \\
       Google Bard & model 7 & model 7 & model 7 \\
       \bottomrule
    \end{tabular}}
    \vspace{0.3cm}
    \quad
    \subfloat[Scenario 3 (large dataset)]{
    \begin{tabular}{lccc}
        \toprule
       LLM & $1^{st}$ try & $2^{nd}$ try & $3^{rd}$ try \\
       \midrule
       ChatGPT 3.5 & model 14 & model 20 & model 20 \\
       ChatGPT 4.0 & model 14 & model 14 & model 14 \\
       Google Bard & model 14 & model 14 & model 14 \\
       \bottomrule
    \end{tabular}}
    \caption{LLM model selection results}
\end{table*}

Unlike the participants who only took the survey once, LLMs have the capacity to answer the same question with multiple responses. This can be achieved by either selecting ``regenerate response" or by starting a new conversation with the identical prompt. For each scenario, we repeated the prompt three times. Table 5 summarizes the ``best" models selected by each LLM in each scenario.

In two out of three scenarios, the LLMs' response patterns closely resembled those of the human participants. In scenario 1, the LLMs' responses varied the most. Responses were evenly split between models 2 and 6, with one mention of model 4. Human preferences paralleled this, with roughly equal preferences for models 2 and 6 and a lesser preference for model 4. In scenario 2, both LLMs and humans had the highest amount of agreements. The vast majority agreed that model 7 is the best model, as this is a relatively easy case. However, in scenario 3, while LLMs gravitated towards model 14, human responses had more variance. The reason for this difference is unknown.

When asked to state model selection reasons, different LLMs also showed varying preferences for parsimony. For instance, in scenario 3, ChatGPT 3.5 stated, ``The preference for simpler models (14 hidden states) make it a reasonable choice. The preference for simpler models is rooted in the principle of Occam's razor, which states that among competing hypotheses or models, the simplest one should be preferred until evidence suggests otherwise." In comparison, instead of simply favoring simpler models, ChatGPT 4.0 stated, ``It's also important to consider the trade-off between model simplicity and the model's ability to capture complex structures in the data." These responses mirrored the humans' responses, where different participants showed different attitudes towards parsimony.

Moreover, similar to 30.3\% of the participants, all LLMs recognized that model selection may be context-dependent. For instance, ChatGPT 4.0 commented, ``This [model selection] also largely depends on the specific context and what you prioritize in model performance. You may consider factors like interpretability, computation cost, and the consequences of misclassification when making a final decision." Similarly, Google Bard said, ``I would also consider the following factors when selecting the best model: the domain knowledge of the problem, the computational resources available, and the time constraints."

Like their human counterparts, all LLMs showed inconsistencies in their answers. Each provided different answers to the same prompt in at least one scenario. We prompted them to rate the importance of the four factors shown in Table 4(a), in the same way we asked the human participants to rate these factors (see Appendix I for an example prompt). Using the method detailed in section 4.2, we derived a consistency score for each LLM. Both ChatGPT 3.5 and Google Bard scored 3 (slightly inconsistent), while ChatGPT 4.0 scored 7 (inconsistent).

In summary, the LLMs mirrored human behavior in model selection. Both the LLMs and the humans expressed varying preferences for parsimony, recognized that model selection may be context-dependent, and showed notable inconsistencies in their decision processes.

\section{Conclusion}

The results highlight that subjectivity can greatly influence the seemingly objective model selection process, especially when different criteria and metrics disagree. Furthermore, a large percentage of human participants and all the LLMs are inconsistent in their model selection preferences, with no trends seen in experience, background, or LLM versions.

Researchers and practitioners must recognize their inherent subjectivity and potential inconsistencies. High levels of subjectivity and inconsistencies cast doubts on the assertions made in ML research and compromise the robustness of ML models. When leveraging tools like LLMs to aid model selection, it is equally important to be cognizant of the limitations and inconsistencies inherent in these tools. 

To improve the repeatability and reproducibility of ML studies, the ML community should standardize the reporting of subjective choices in model selection. Our study revealed several sources of subjectivity. First, both the participants and the LLMs had very different views on how parsimonious models should be. The preferred degree of parsimony should be reported and justified in application contexts. This preference should also guide which ICs are considered more important because different ICs quantify different trade-offs between parsimony and accuracy. 

Another source of subjectivity is how different dataset sizes should influence model selection. Multiple studies agree that AIC (and other ICs that emphasize accuracy) is preferred for small datasets because underfitting is more likely a concern, while BIC (and other ICs that emphasize parsimony) is preferred for large datasets because overfitting is more likely a concern \cite{costa2010model, dziak2020sensitivity}. Any model selection preferences that violate this recommendation should be reported and justified.

If researchers and practitioners run cross-validation tests in addition to information criteria, then cross-validation metrics should also be reported, since it is a major source of subjectivity. Furthermore, if many metrics disagree, researchers and practitioners should consider whether the model should be trained in the first place. For instance, if many different ICs and cross-validation metrics suggest HMMs with widely-varying hidden states, researchers should consider whether an HMM best describes the data.

\subsection{Limitations and future work}

One limitation of our study is that it is unknown whether multiple ICs and cross-validation metrics are used in practice. Some studies use BIC alone to select models without discussing alternative ICs \cite{le1999monitoring, wang2016hidden}. 

Another limitation is that our survey did not provide any context about the dataset. As ten participants and all three LLMs commented, context might influence their model selection preferences. Future studies should examine how different contexts affect model selection. However, providing more context might introduce more confounding variables. For example, had participants been informed that the models were trained on StarCraft II data, their prior knowledge of StarCraft II might have influenced their preferences. Hence, future studies will need careful experimental designs to effectively disentangle these confounding influences.

This study revealed that many participants were either unfamiliar with model selection guidelines in the literature or chose to ignore them, as evidenced by varying opinions regarding dataset sizes. Future studies should examine whether or when researchers and practitioners are inclined to follow given technical guidelines and why. Another direction for future studies is whether a data visualization tool could make the model selection process more transparent. Five participants (15.2\%) mentioned the information presented in the survey seemed overwhelming, and some visualizations of the data might help. 

Ultimately, a better understanding of the researcher degrees of freedom in model selection will inform a more standardized model selection process, which will help to improve the repeatability and reproducibility of various ML studies and ultimately lead to more confidence in various ML models deployed in the real world.  

\bibliography{aaai24}

\clearpage
\onecolumn
\appendix

\section{Theoretical and practical motivation of different ICs}

AIC aims to select the candidate model expected to have the lowest error in predicting future data taken from the same process as training data \cite{dziak2020sensitivity}. AIC estimates the relative Kullback-Leibler divergence of the likelihood function specified by a model from the likelihood function of the unknown true process that generated the data \cite{dziak2020sensitivity}. 

Since Akaike first introduced AIC in 1974, researchers have proposed various modifications of AIC. Some found that changing the penalty term from $2n_p$ to $3n_p$ improves simulation performance in some settings, so they proposed AIC3 \cite{andrews2003comparison}. Other researchers found that when the sample size is small, or when the number of parameters is a moderate to large fraction of the sample size, AIC tends to overfit \cite{hurvich1989regression}. AICc tries to fix this tendency by applying a slightly heavier penalty based on dataset size $T$ and the number of parameters \cite{hurvich1989regression}. AICu, another alternative, is developed as an unbiased estimator of Kullback-Leibler information \cite{mcquarrie1997model}. Researchers found that AICu outperforms AICc for moderate to large sample sizes except when the true model is of infinite order \cite{mcquarrie1997model}.

CAIC ("Consistent AIC") is yet another variation. It is motivated by making AIC consistent \cite{bozdogan1987model}. Consistent information criteria select the correct model with probability approaching 1 in large samples when the true model is of finite dimension and belongs to the set of candidate models \cite{bozdogan1987model}.

BIC has a different theoretical motivation. BIC is derived from the Bayes theorem and aims to select the candidate model that is a posteriori most probable \cite{neath2012bayesian}. It is one of the most widely known and pervasively used tools in statistical model selection \cite{neath2012bayesian}. Compared to AIC, BIC has a heavier penalty weight and focuses more on parsimony. Researchers have also proposed various modifications of BIC. For example, ABIC is a sample-size-adjusted version of BIC. The penalty term of ABIC is derived from model selection for autoregressive time series models \cite{sclove1987application}.

\section{Complete IC results}

The minimum value of each column is highlighted in green, and the models with the minimum values are selected as ``best" models. 

\begin{figure}[!htb]
    \centering
    \includegraphics{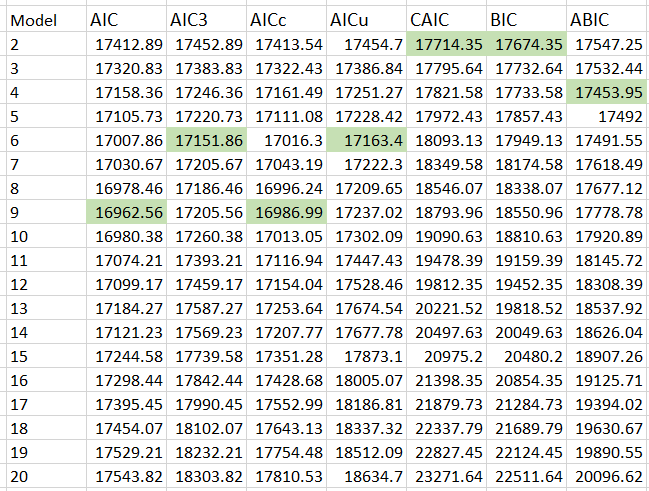}
    \caption{IC results for scenario 1}
\end{figure}

\begin{figure}
    \centering
    \includegraphics{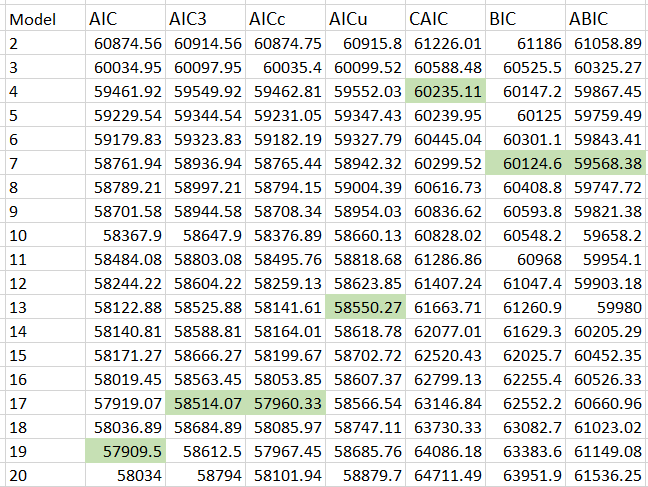}
    \caption{IC results for scenario 2}
\end{figure}

\begin{figure}
    \centering
    \includegraphics{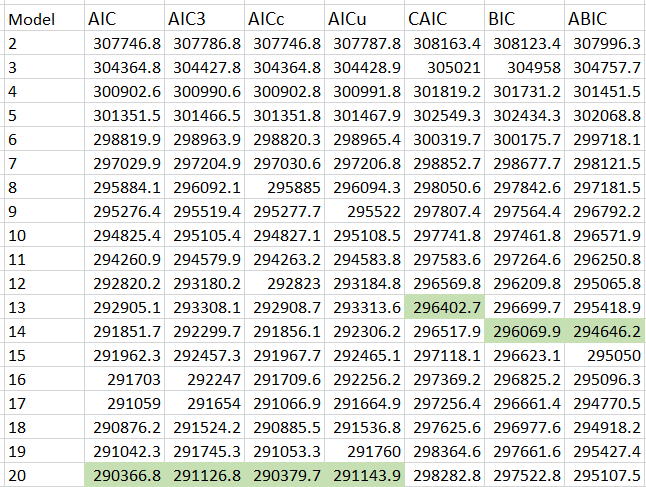}
    \caption{IC results for scenario 3}
\end{figure}

\pagebreak

\section{Cross-validation results}
Here we present the cross-validation results. The second column, “$-2(\log L)$ term average,” takes the average of the negative loglikelihood term across the 5 or 10 folds. Lower  negative loglikelihood term values suggest that the model has a higher likelihood. For example, in scenario 1, model 4 is the model with the “best average.” To determine whether model 4’s  negative loglikelihood term average is significantly better than that of the other model’s, paired T-tests were performed. In scenario 1, the results are insignificant. The fourth column is the standard deviation across the 5 or 10 folds. A lower standard deviation suggests that the model is more consistent. By this standard, in scenario 1, model 6 is “most consistent.” Lastly, for each model, the largest loglikelihood term value among the 5 folds is displayed as the “worst case” in the last column. In scenario 1, model 4 has the “best worst case.” 

\begin{table}[!htb]
    \centering
    \begin{tabular}{lccccc}
    \toprule
    Model & $-2(\log L)$ term average & Paired T-test & $-2(\log L)$ term SD & Worst case\\
    \midrule
    2 & 3604.81 & N/A & 277.59 & 4000.50\\
    4 & 3661.31 & Not significant between models 2 and 4 & 256.96 & 3973.74\\
    6 & 3788.68 & Not significant between models 2 and 6 & 247.73 & 4042.62\\
    9 & 3806.31 & Not significant between models 2 and 9 & 293.81 & 4206.82 \\
    \bottomrule
    \end{tabular}
    \caption{Cross-validation results for scenario 1}
\end{table}

\begin{table}[!htb]
    \centering
    \begin{tabular}{lccccc}
    \toprule
    Model & $-2(\log L)$ term average & Paired T-test & $-2(\log L)$ term SD & Worst case\\
    \midrule
   4 & 12672.53 & N/A & 710.77 & 13615.59\\
    7 & 12686.57 & Not significant between models 4 and 7 & 617.00 & 13020.98\\
    13 & 12730.73 & Not significant between models 4 and 13 & 830.44 & 13349.28\\
    17 & 12733.16 & Not significant between models 4 and 17 & 690.22 & 13373.54 \\
    19 & 12754.90 & Not significant between models 4 and 19 & 735.32 & 13209.94 \\
    \bottomrule
    \end{tabular}
    \caption{Cross-validation results for scenario 2}
\end{table}

\begin{table}[!htb]
    \centering
    \begin{tabular}{lccccc}
    \toprule
    Model & $-2(\log L)$ term average & Paired T-test & $-2(\log L)$ term SD & Worst case\\
    \midrule
   13 & 32083.97 & Not significant between models 14 and 13 & 1320.71 & 33925.16\\
    14 & 32061.27 & N/A & 1341.95 & 34213.48\\
    20 & 32102.46 & Not significant between models 14 and 20 & 1321.69 & 33866.94\\
    \bottomrule
    \end{tabular}
    \caption{Cross-validation results for scenario 3}
\end{table}

\pagebreak

\section{Model selection study PowerPoint}

\hspace{1cm}

\includegraphics[scale=0.38]{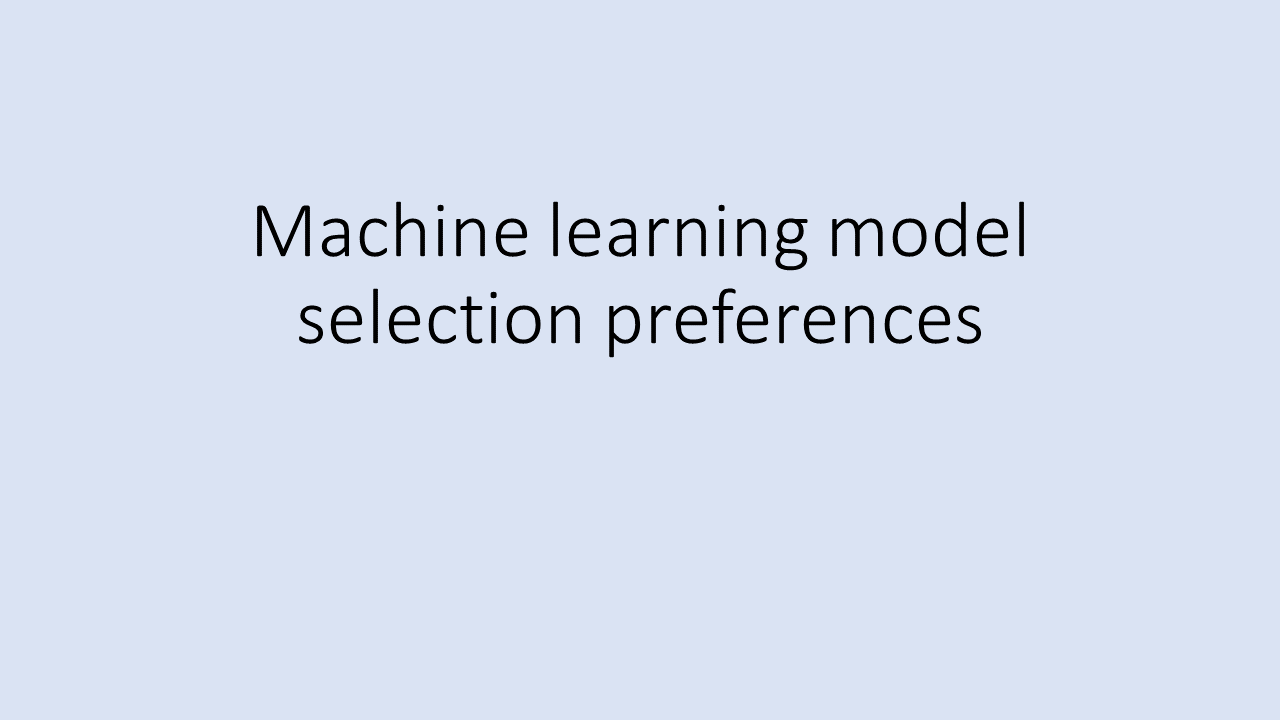}
\vspace{0.1cm}

\includegraphics[scale=0.38]{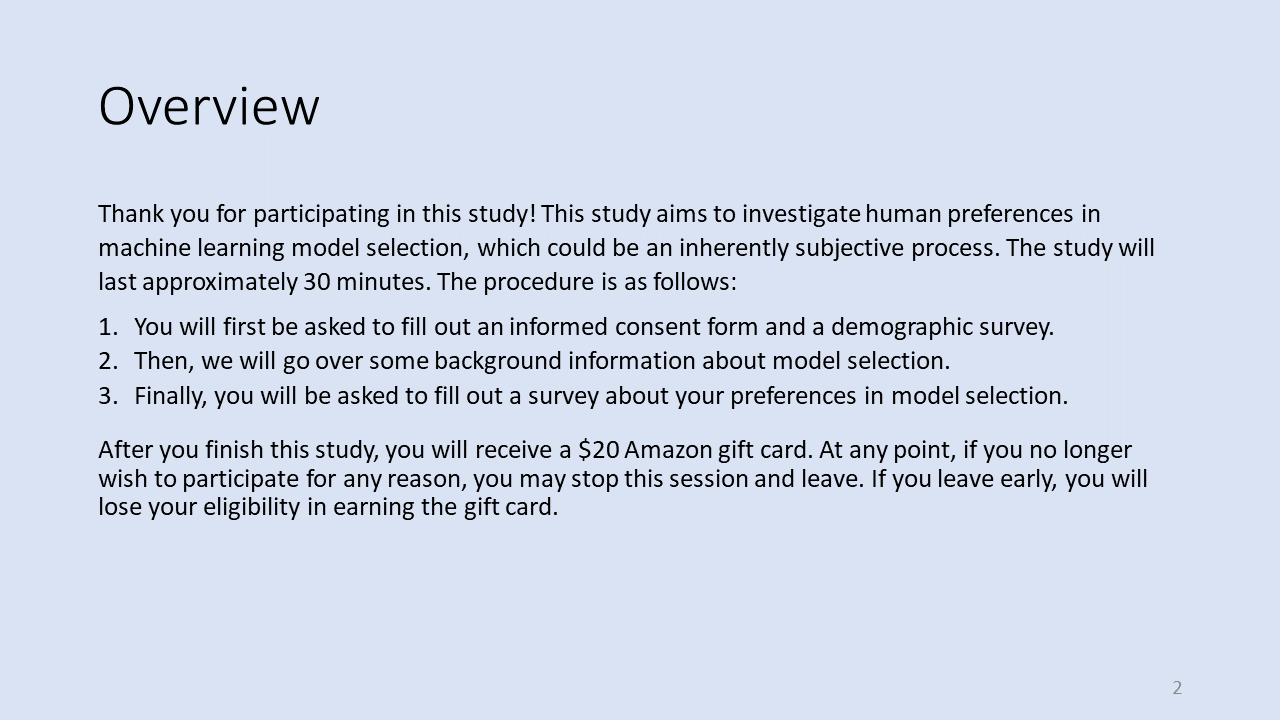}
\vspace{0.1cm}

\includegraphics[scale=0.38]{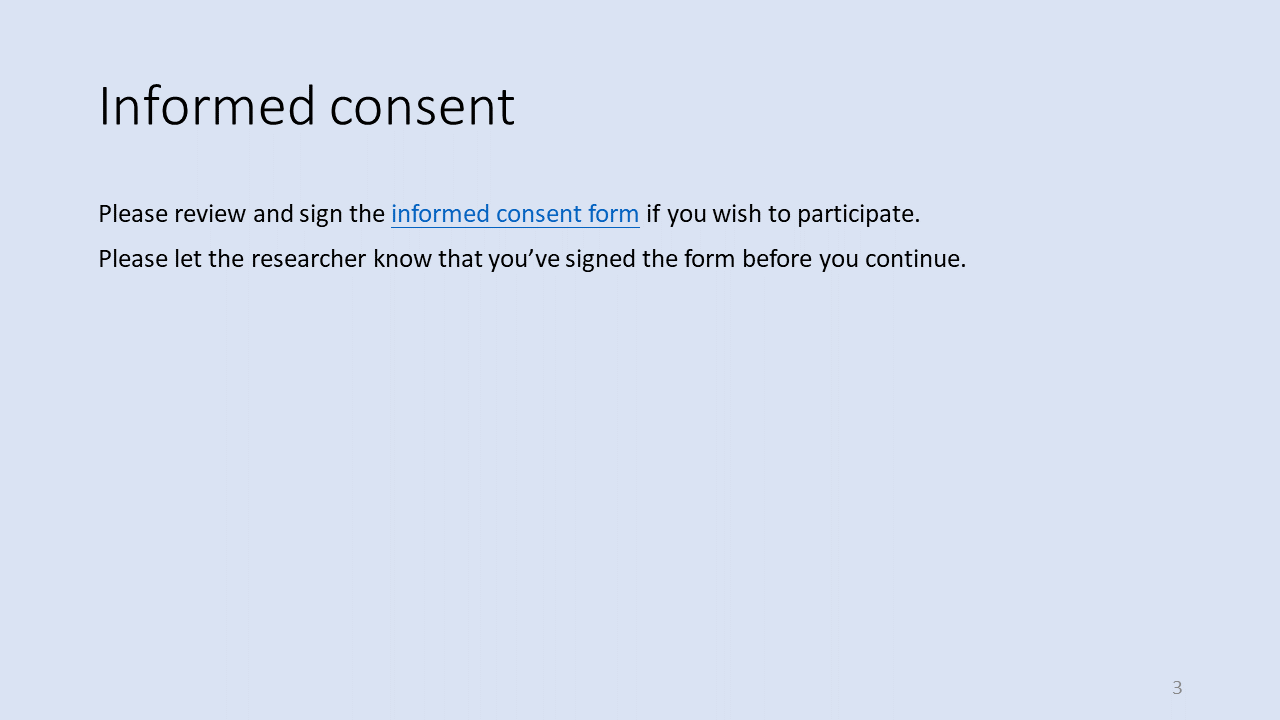}
\vspace{0.1cm}

\includegraphics[scale=0.38]{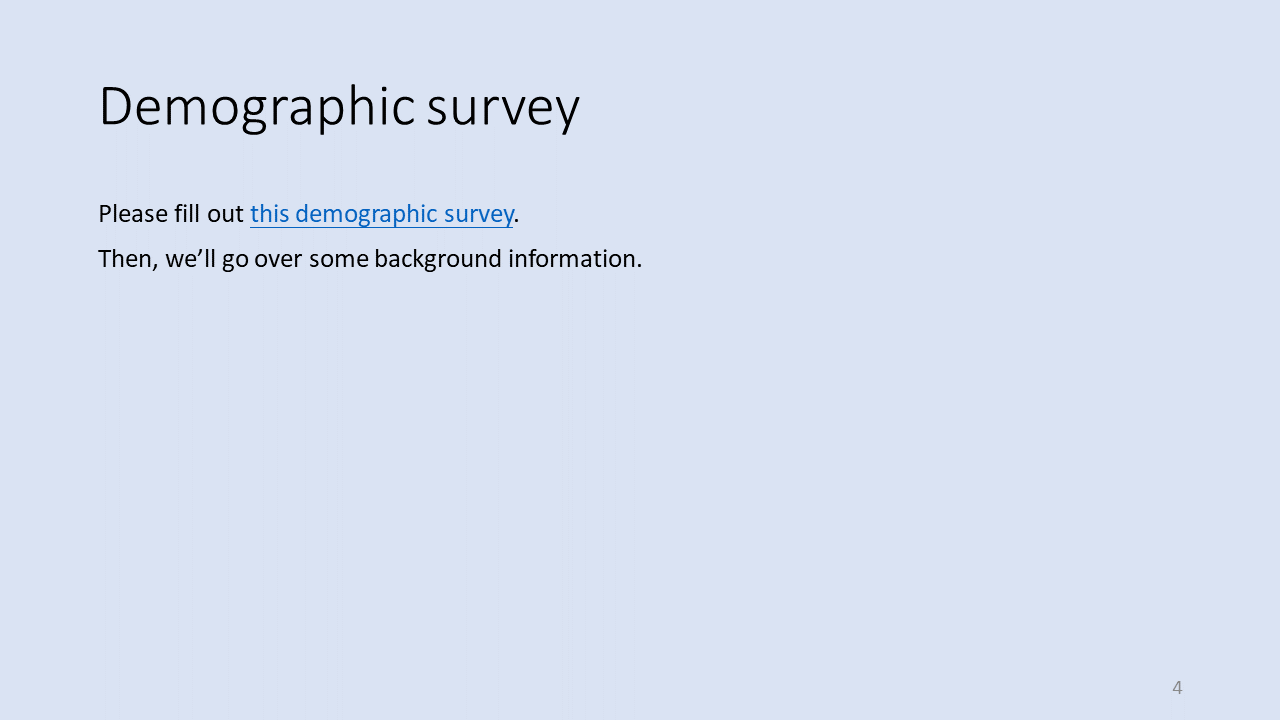}
\vspace{0.1cm}

\includegraphics[scale=0.38]{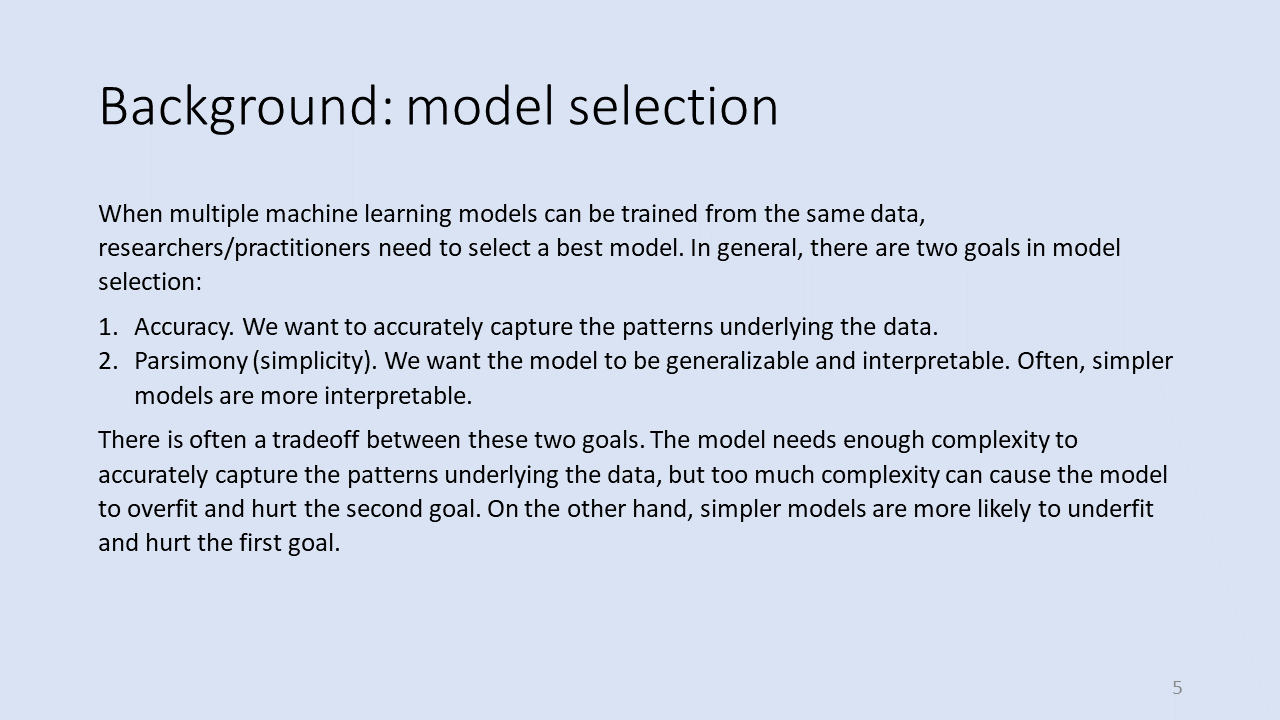}
\vspace{0.1cm}

\includegraphics[scale=0.38]{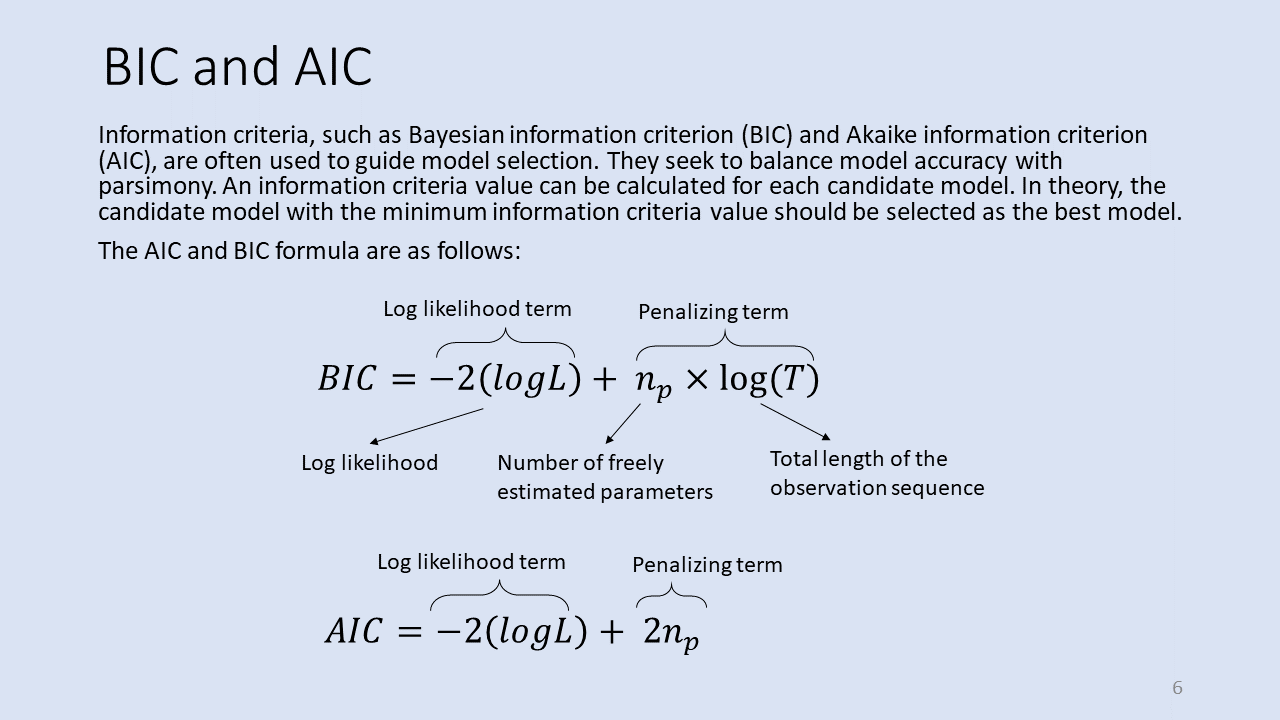}
\vspace{0.1cm}

\includegraphics[scale=0.38]{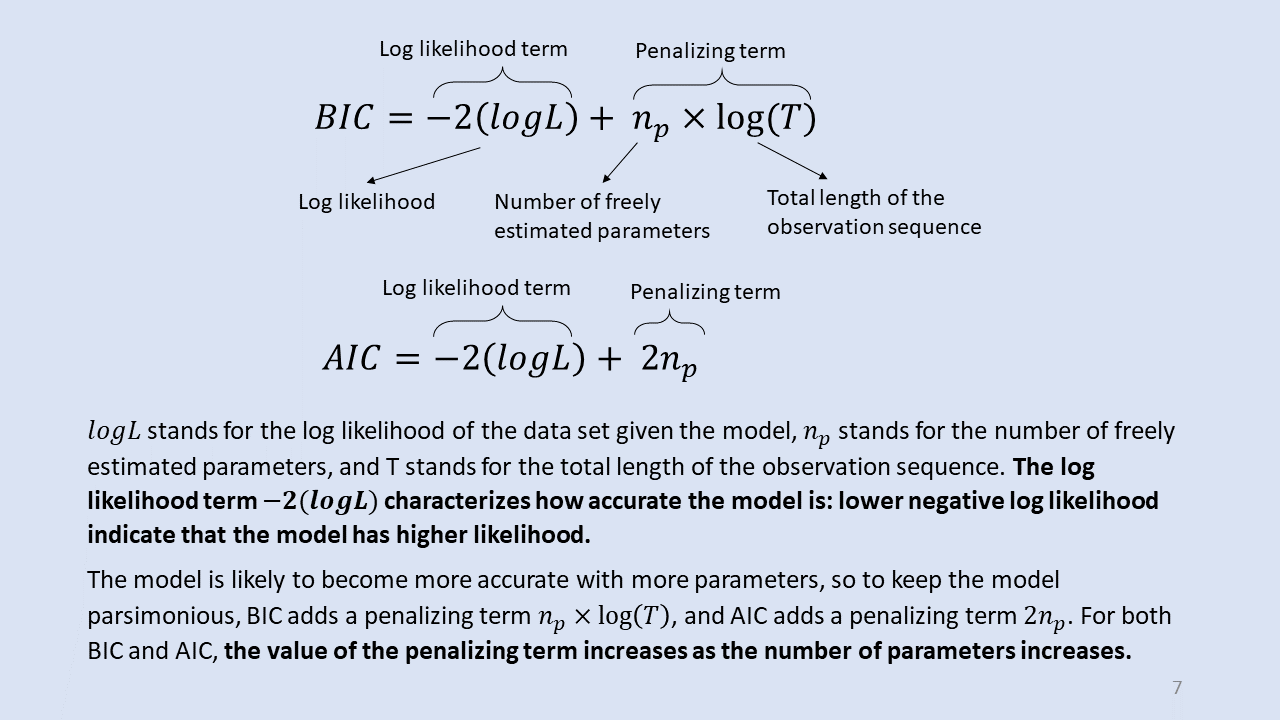}
\vspace{0.1cm}

\includegraphics[scale=0.38]{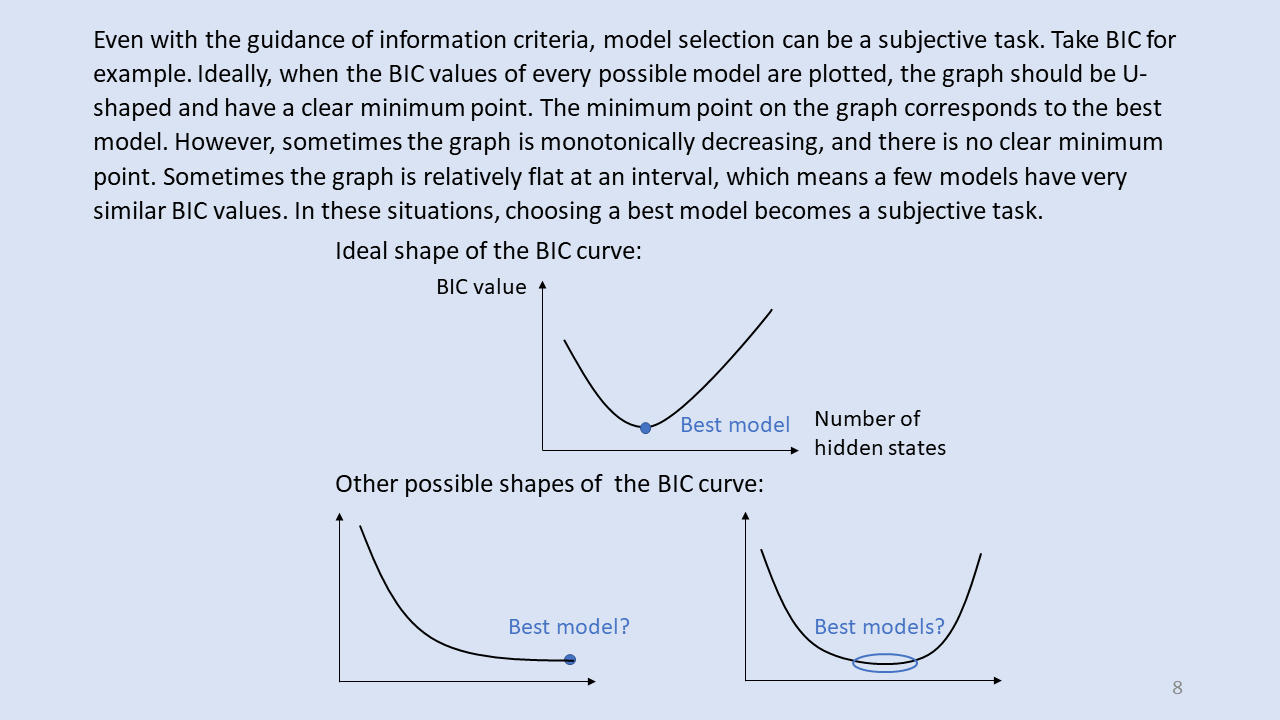}
\vspace{0.1cm}

\includegraphics[scale=0.38]{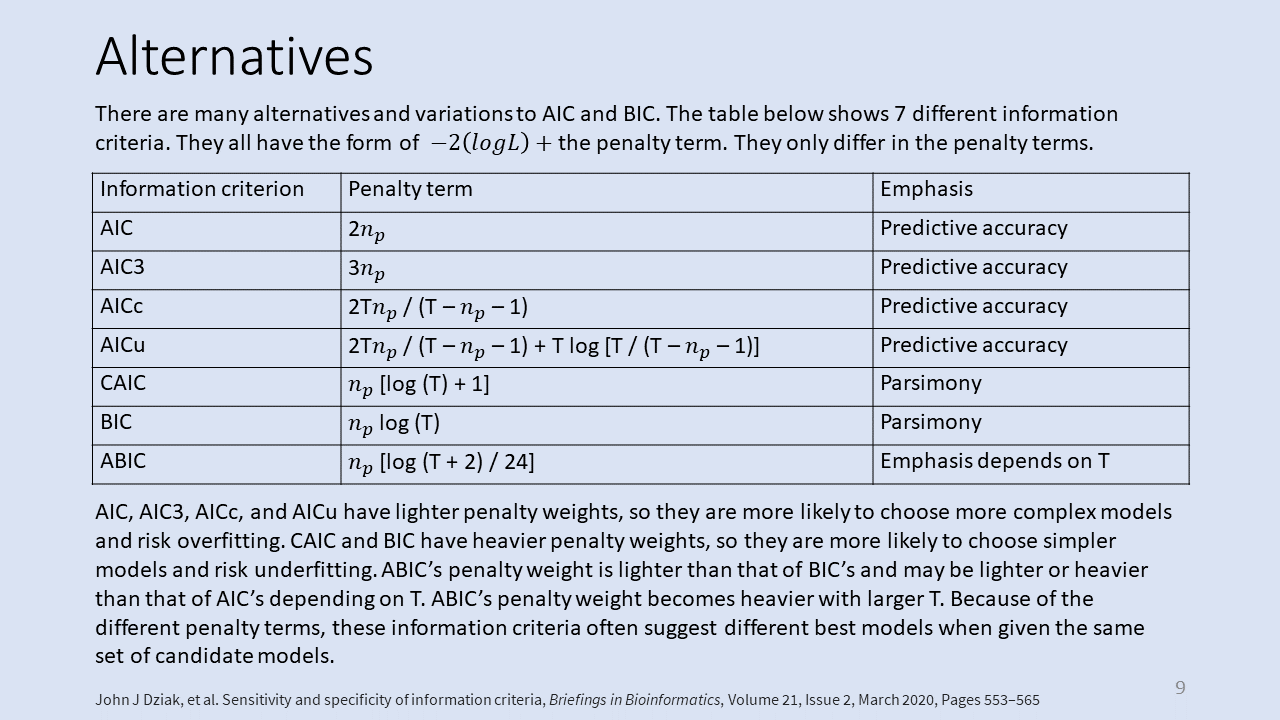}
\vspace{0.1cm}

\includegraphics[scale=0.38]{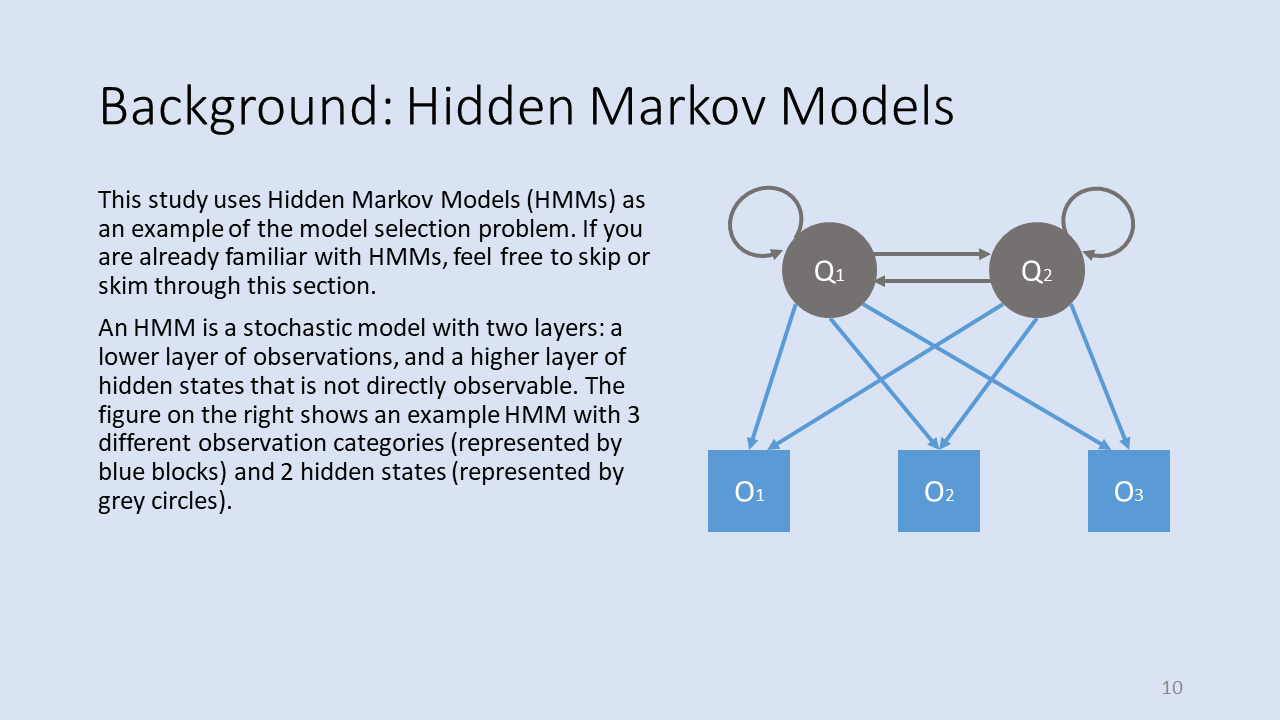}
\vspace{0.1cm}

\includegraphics[scale=0.38]{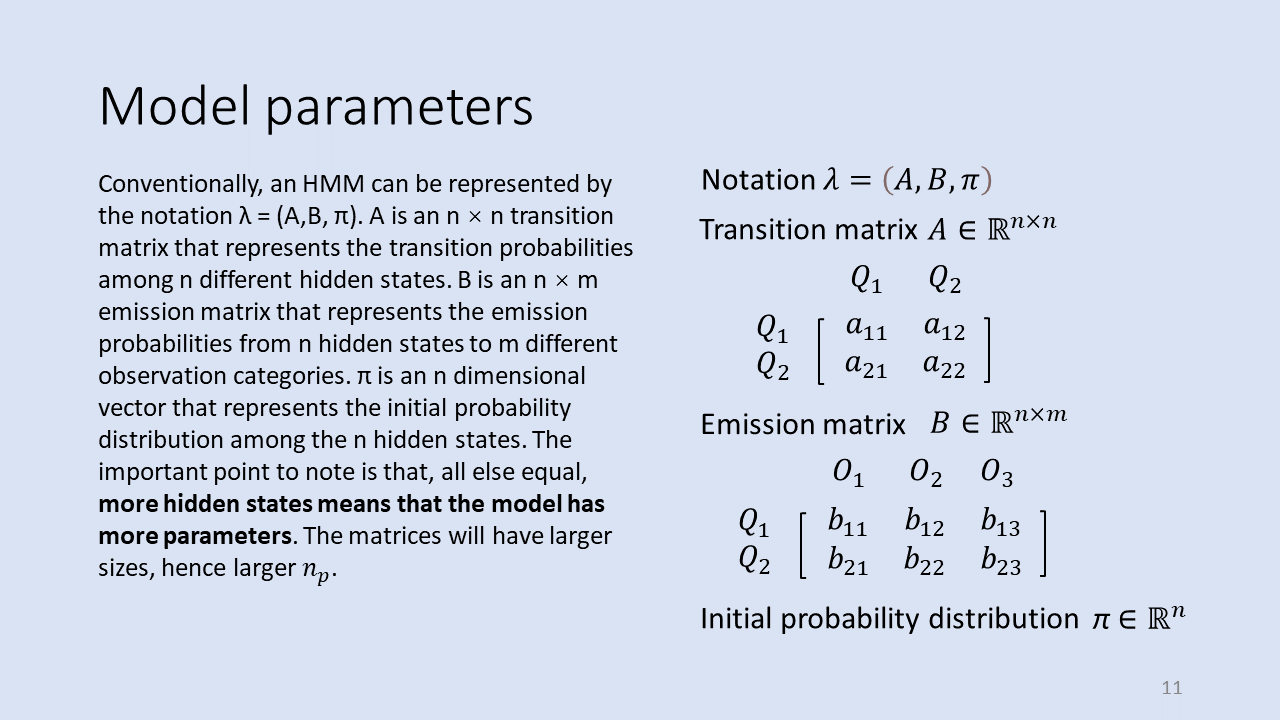}
\vspace{0.1cm}

\includegraphics[scale=0.38]{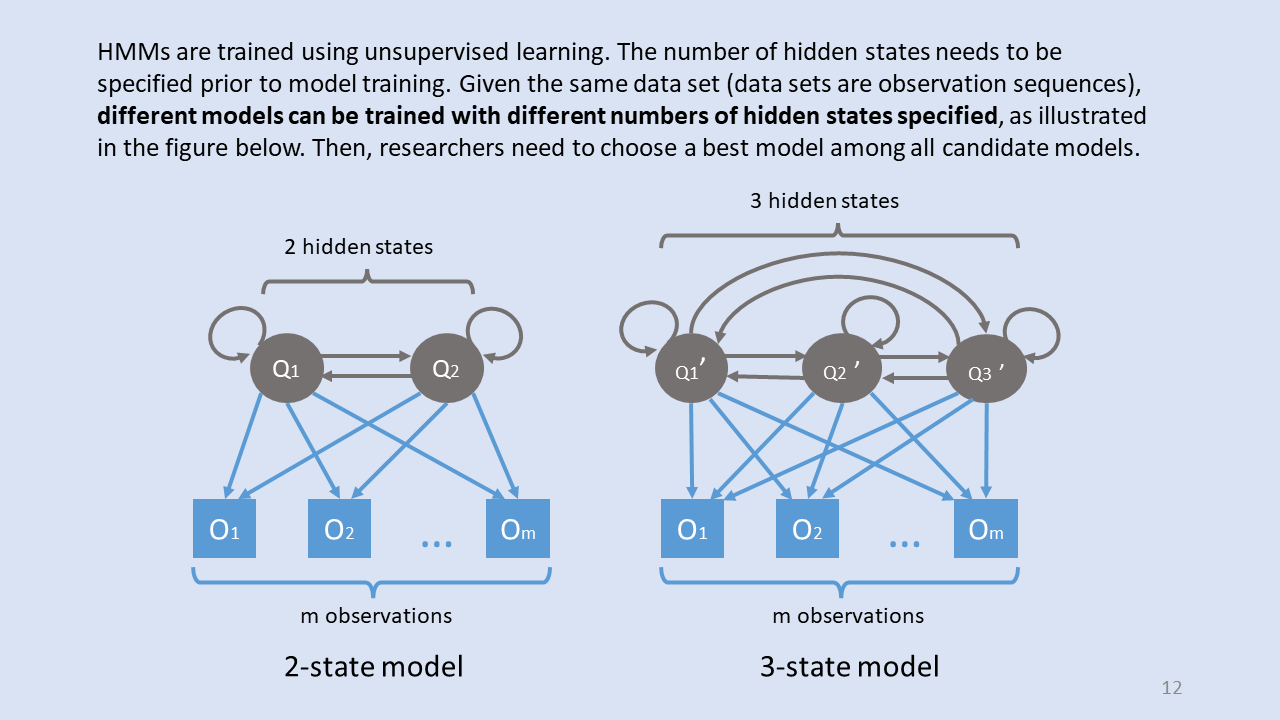}
\vspace{0.1cm}

\includegraphics[scale=0.38]{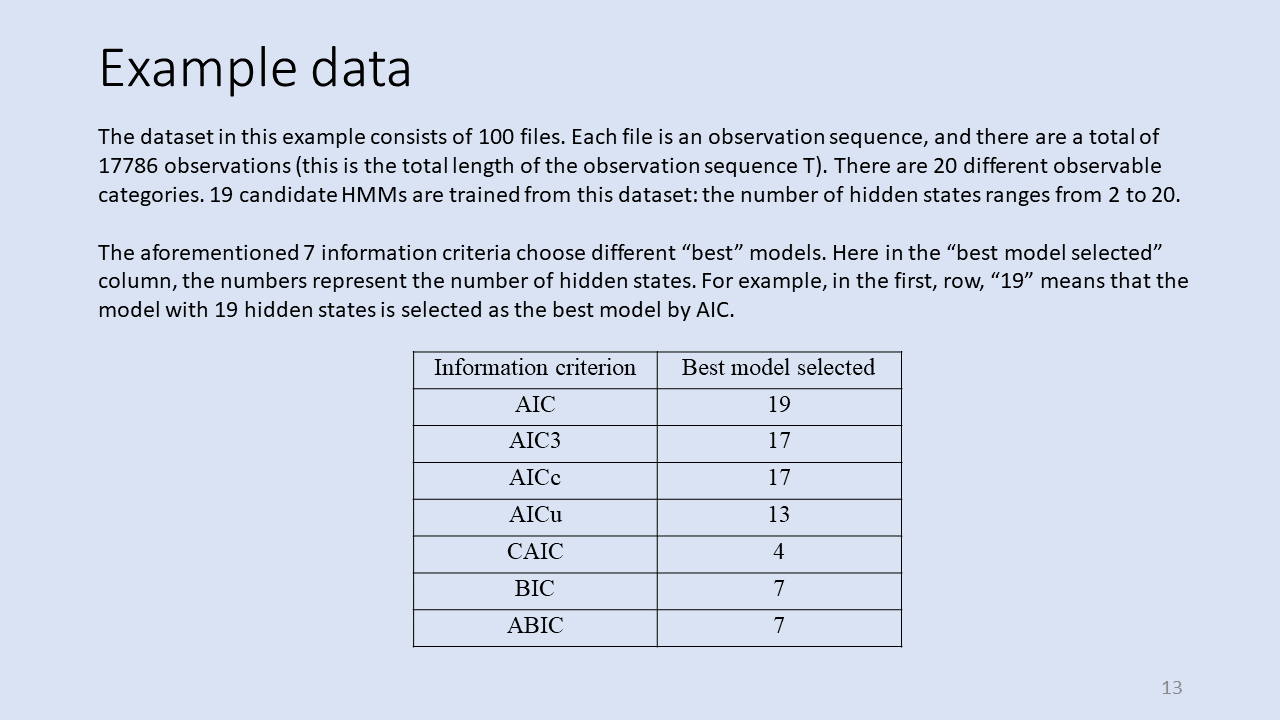}
\vspace{0.1cm}

\includegraphics[scale=0.38]{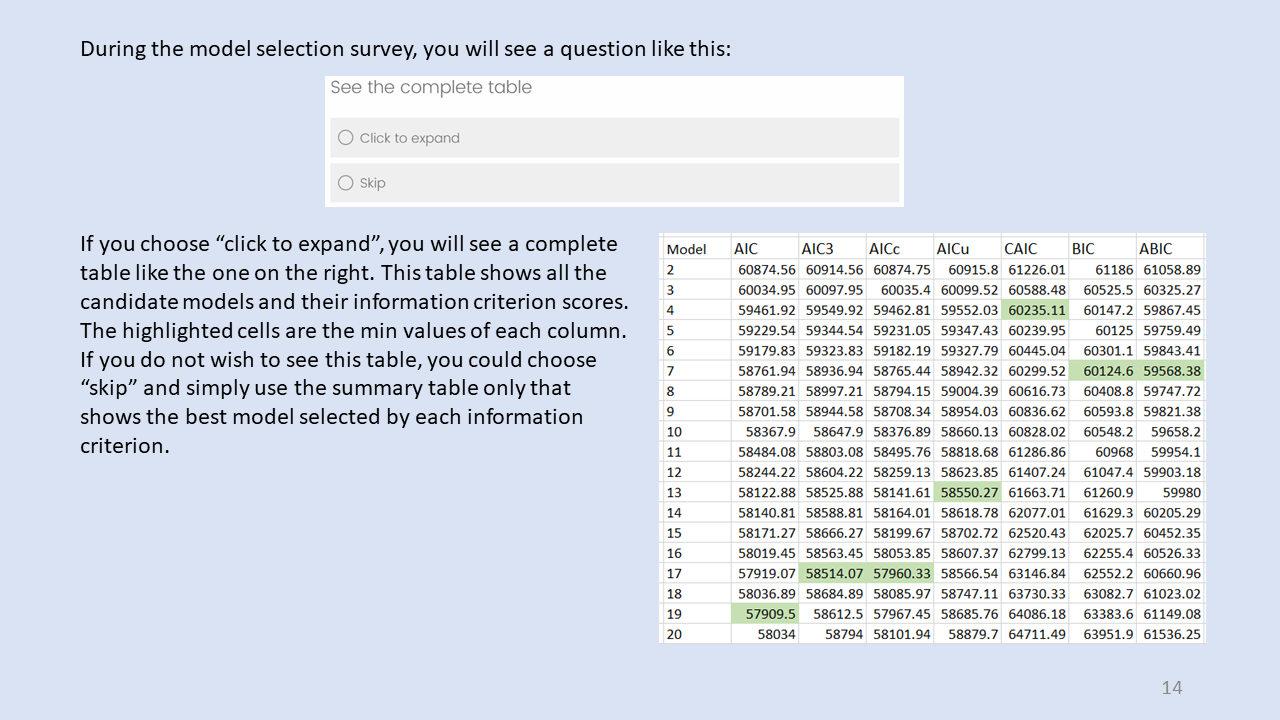}
\vspace{0.1cm}

\includegraphics[scale=0.38]{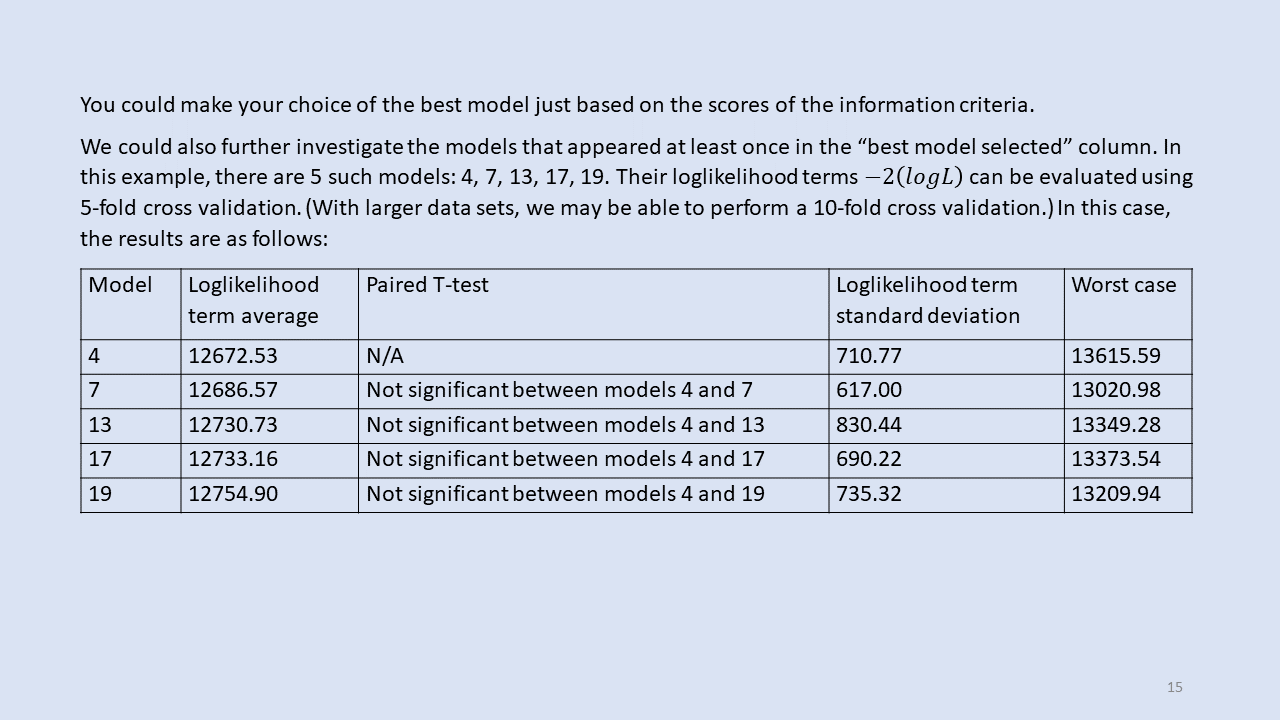}
\vspace{0.1cm}

\includegraphics[scale=0.38]{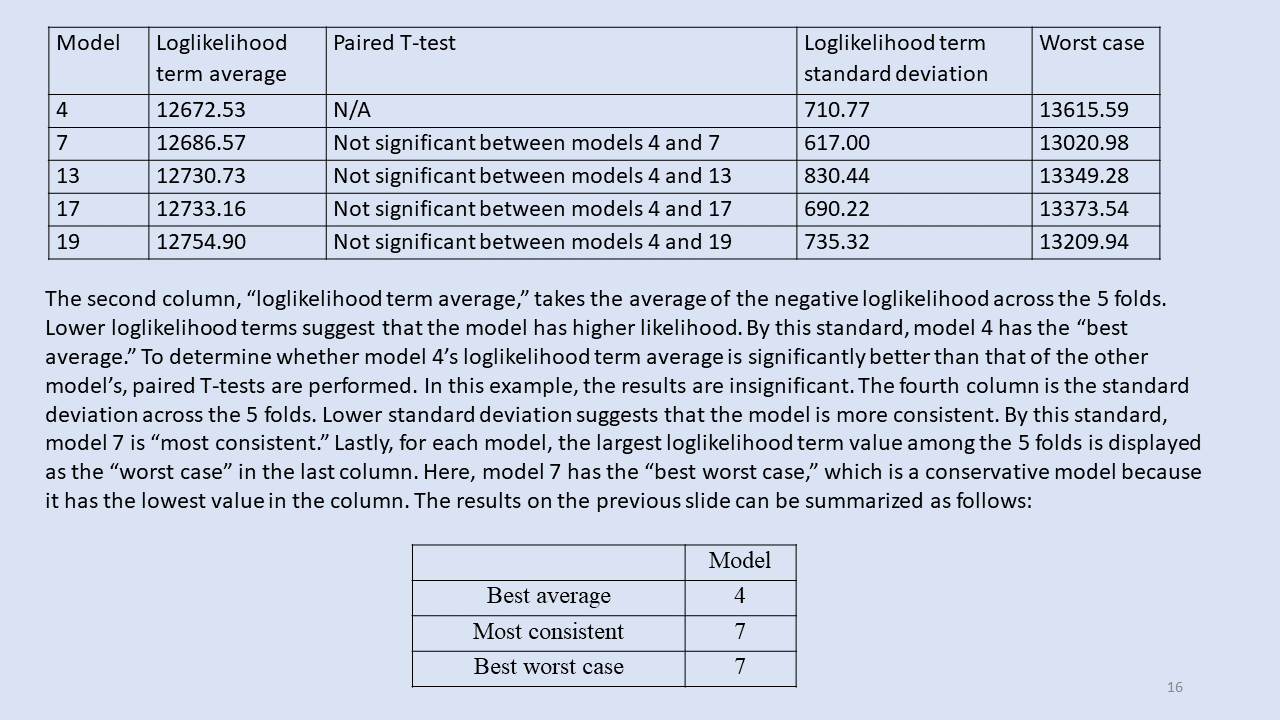}
\vspace{0.1cm}

\includegraphics[scale=0.38]{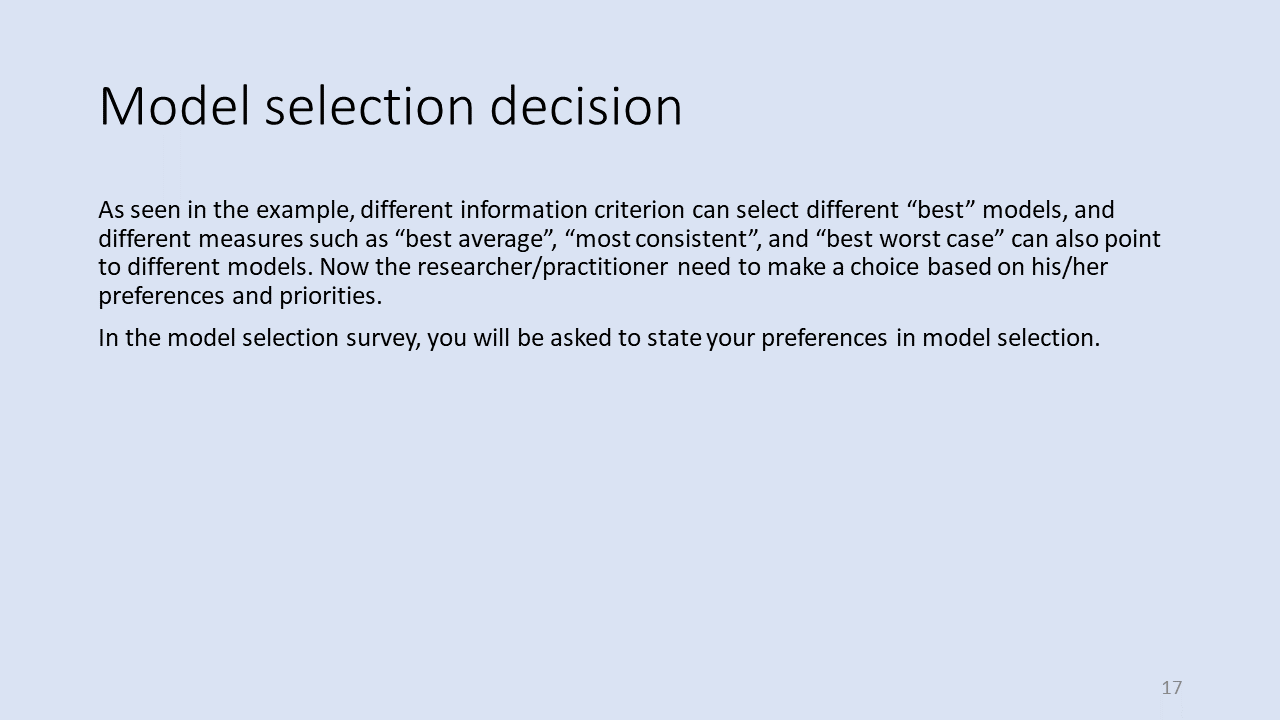}
\vspace{0.1cm}

\includegraphics[scale=0.38]{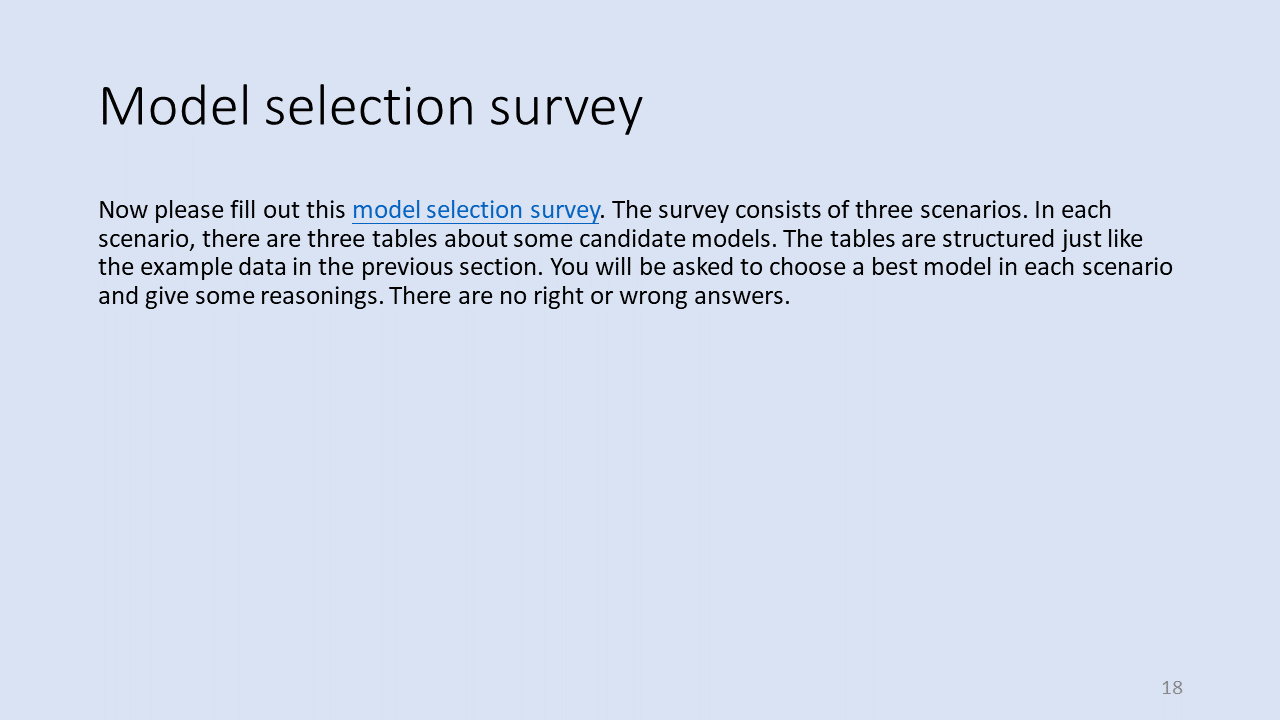}
\vspace{0.1cm}

\pagebreak

\section{Demographics survey}
\begin{figure}[!htb]
    \includegraphics[scale=0.75]{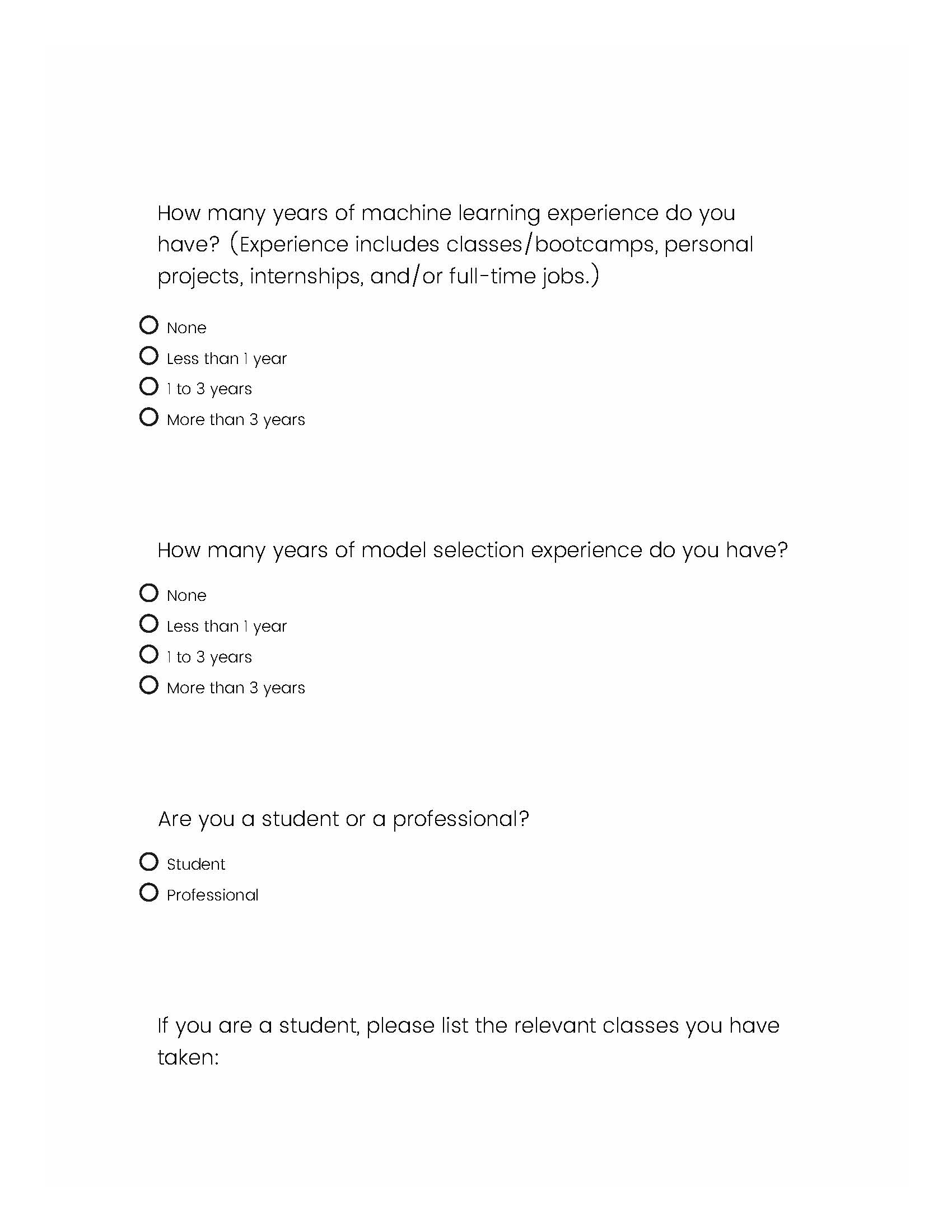}
\end{figure}

\begin{figure}
    \includegraphics[scale=0.75]{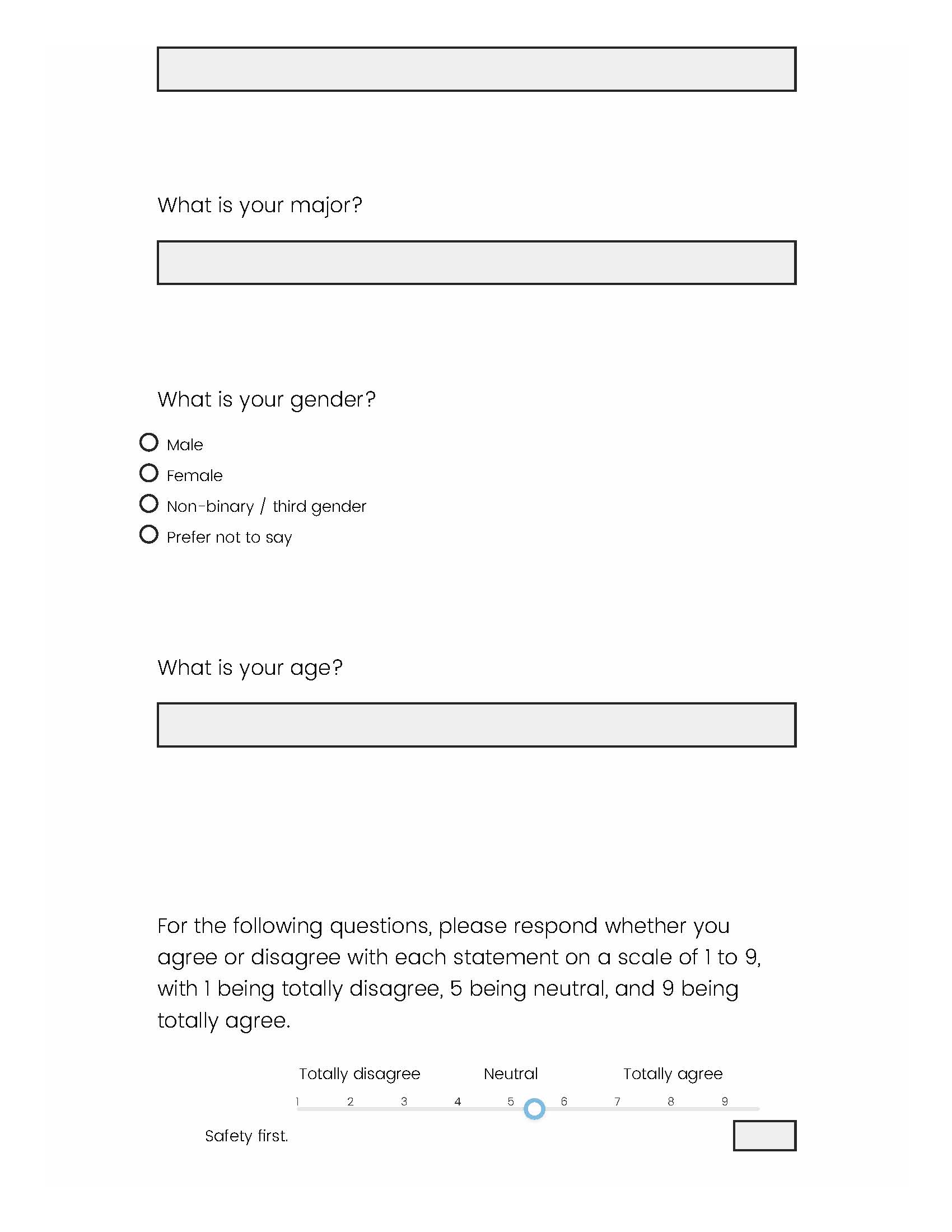}
\end{figure}

\begin{figure}
    \includegraphics[scale=0.75]{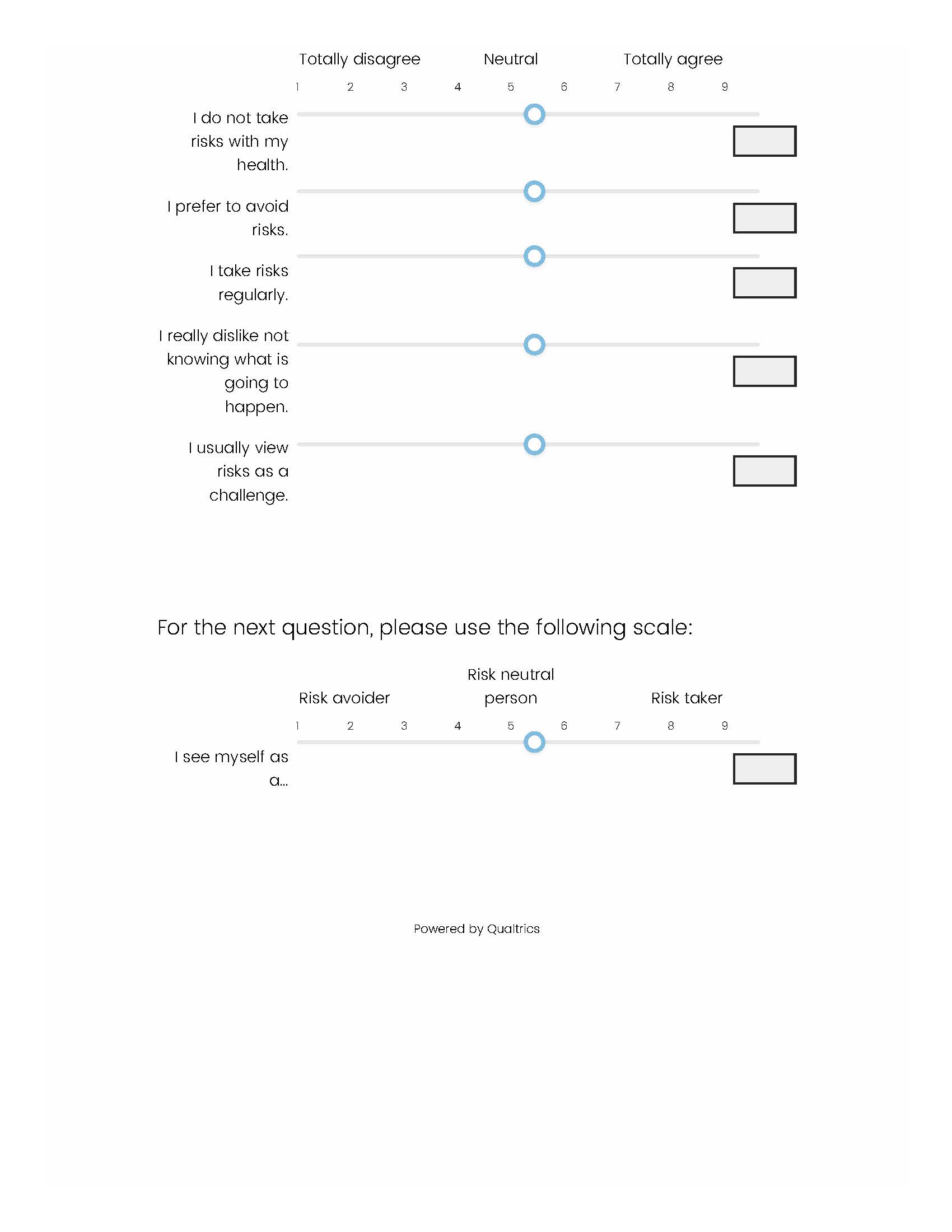}
\end{figure}

\pagebreak

\section{Model selection survey}
    
\begin{figure}[!htb]
    \includegraphics[scale=0.75]{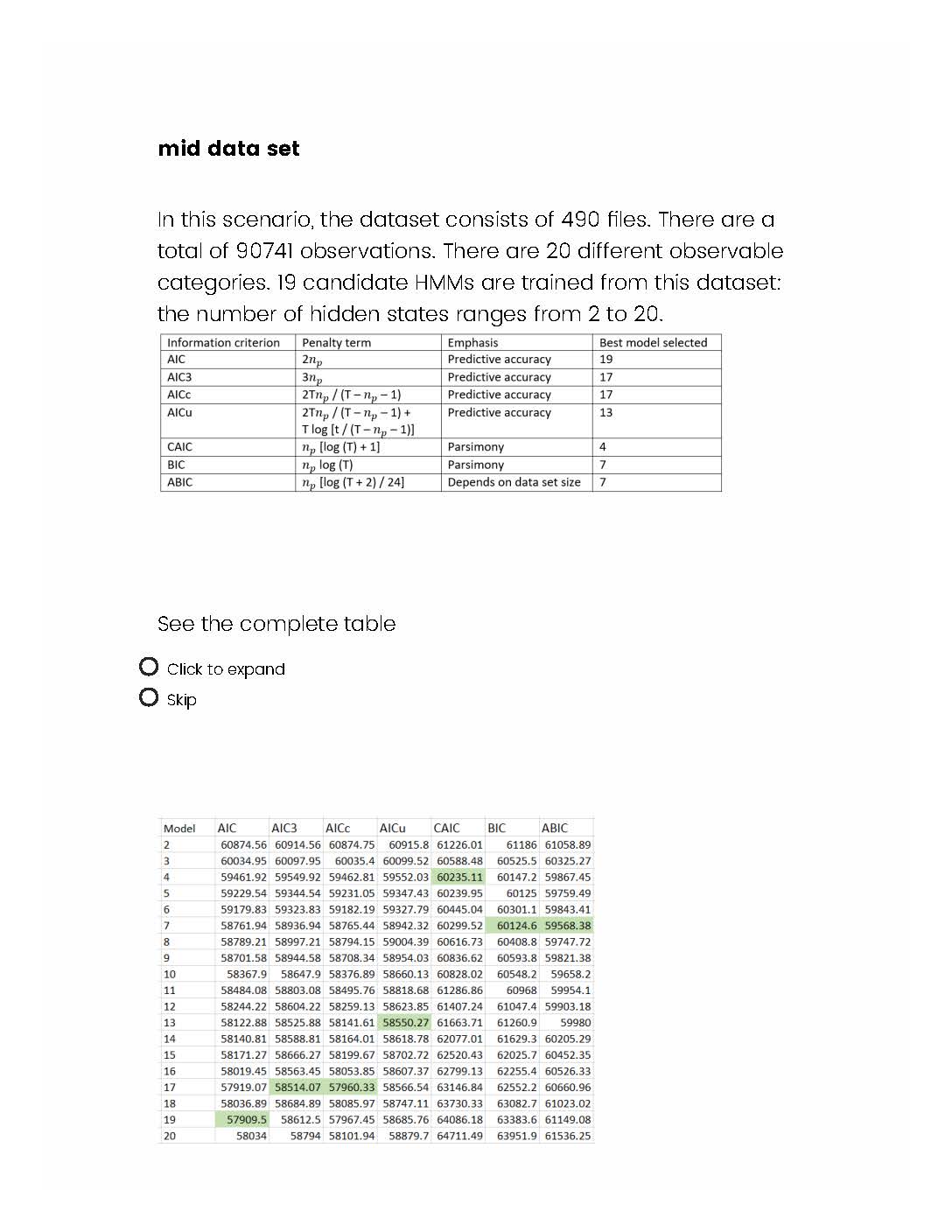}
\end{figure}
\begin{figure}
    \includegraphics[scale=0.75]{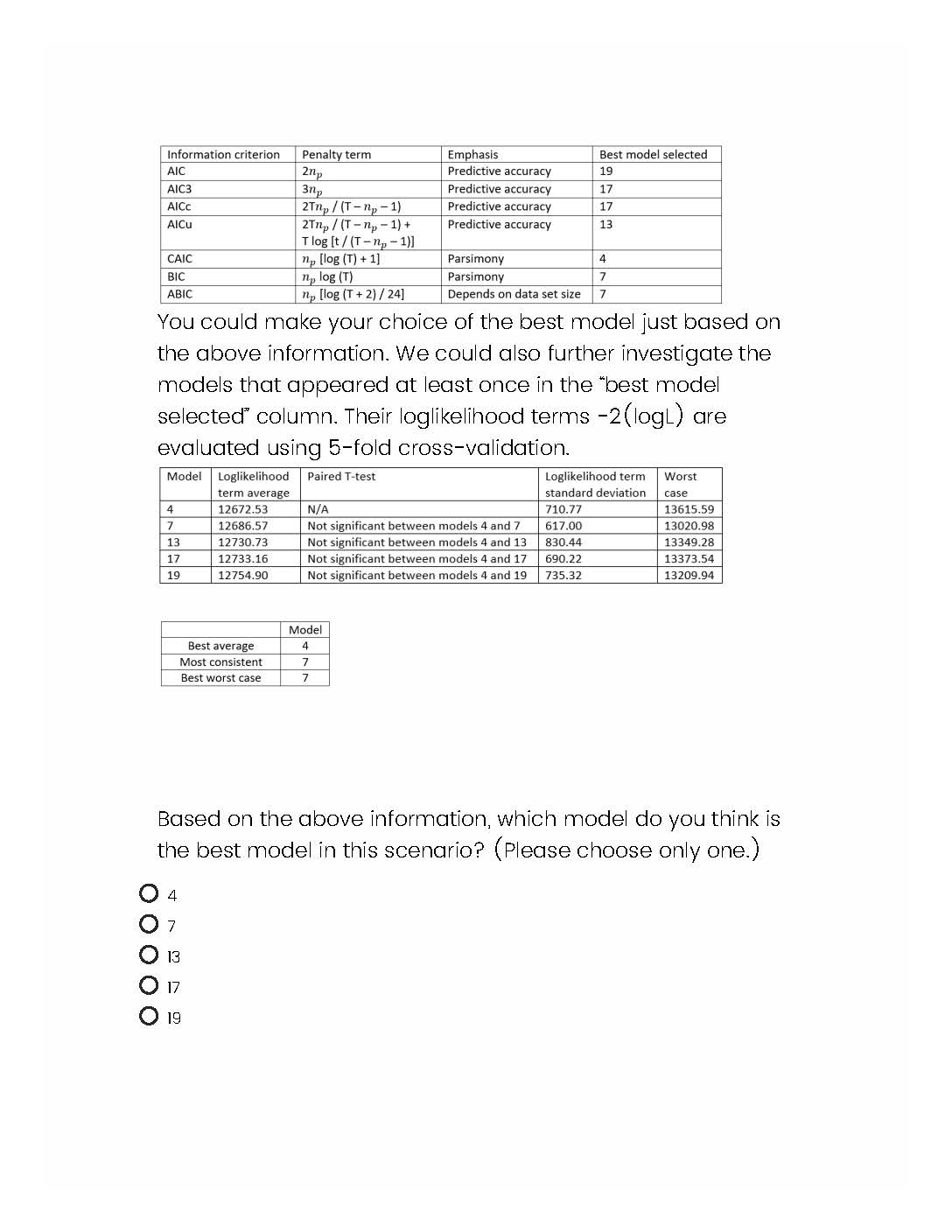}
\end{figure}
\begin{figure}
    \includegraphics[scale=0.75]{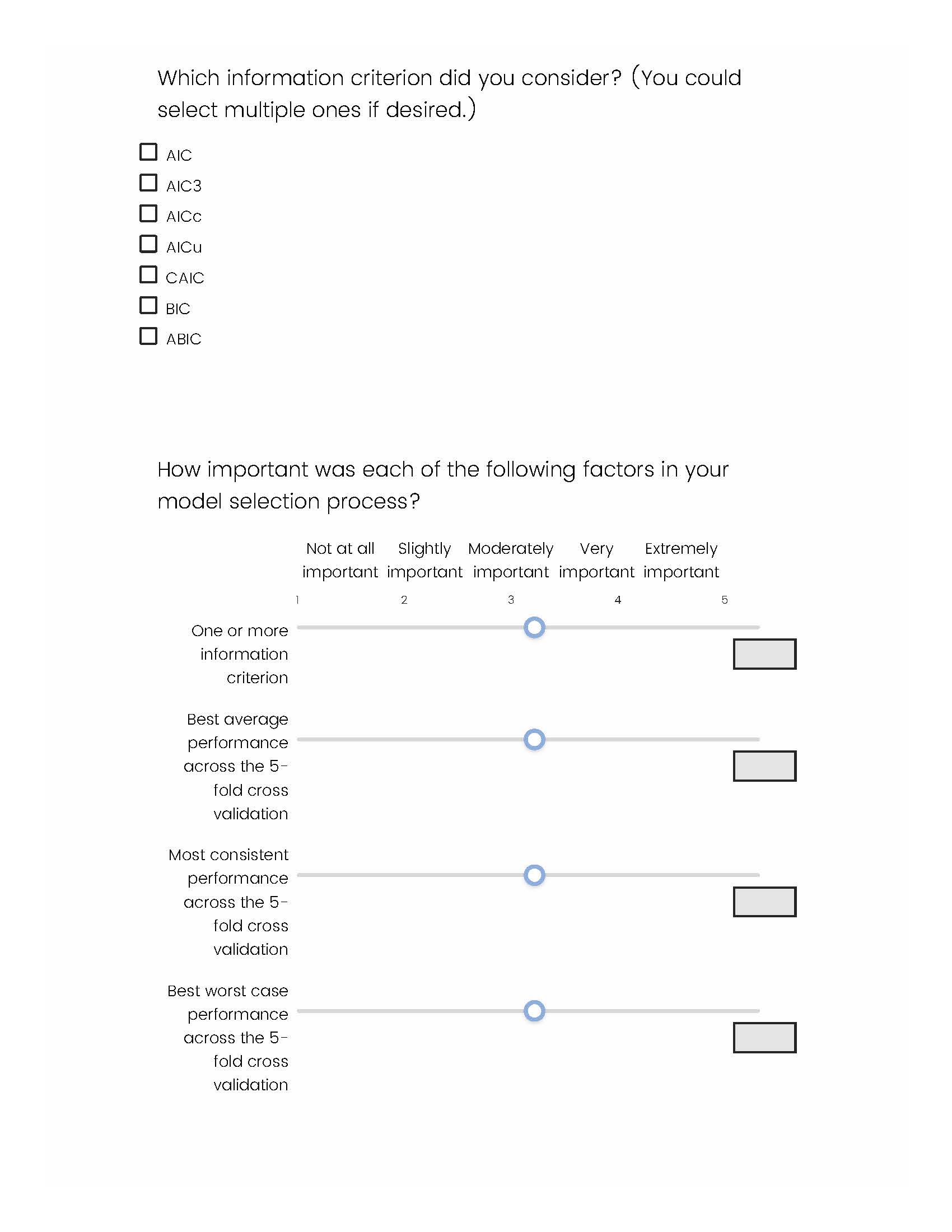}
\end{figure}
\begin{figure}
    \includegraphics[scale=0.75]{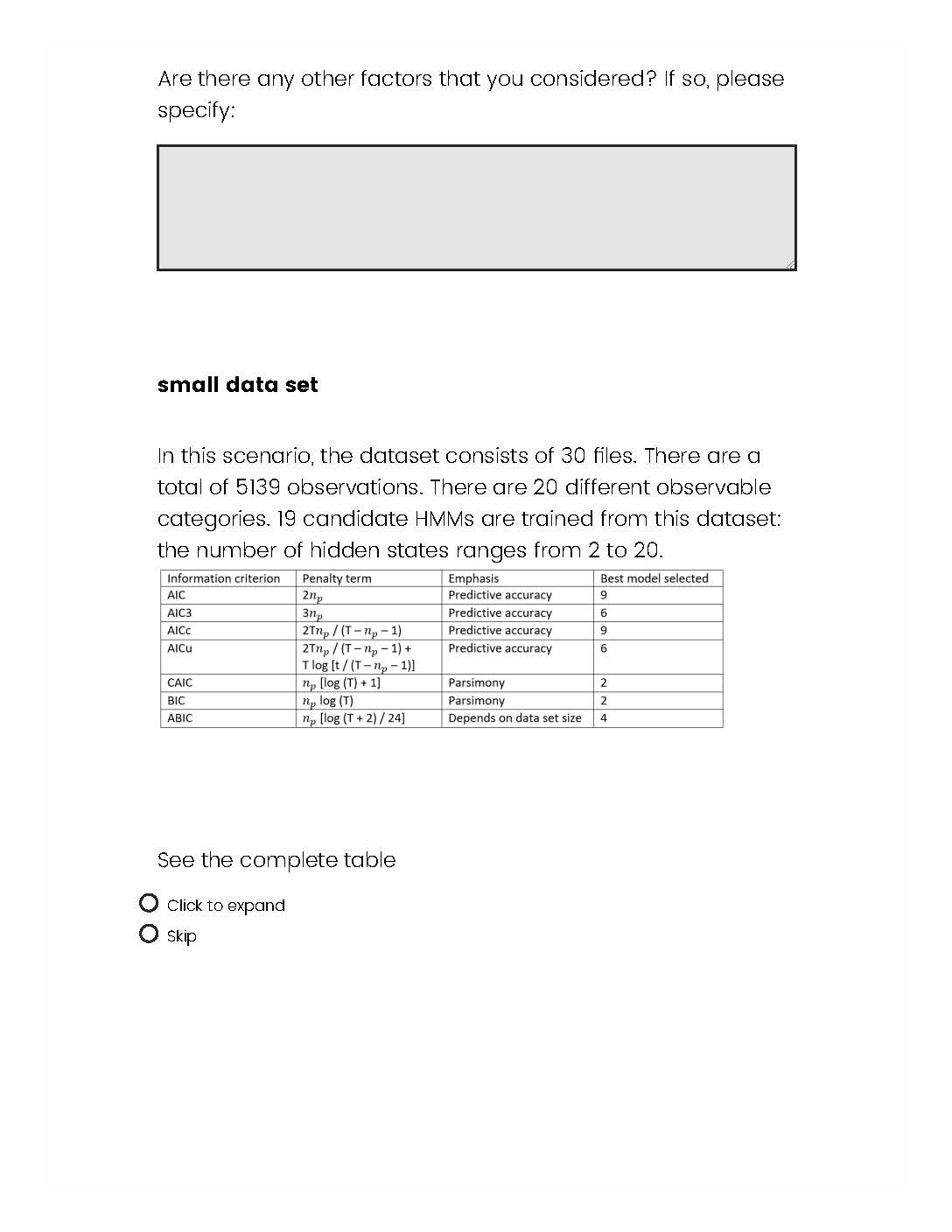}
\end{figure}
\begin{figure}
    \includegraphics[scale=0.75]{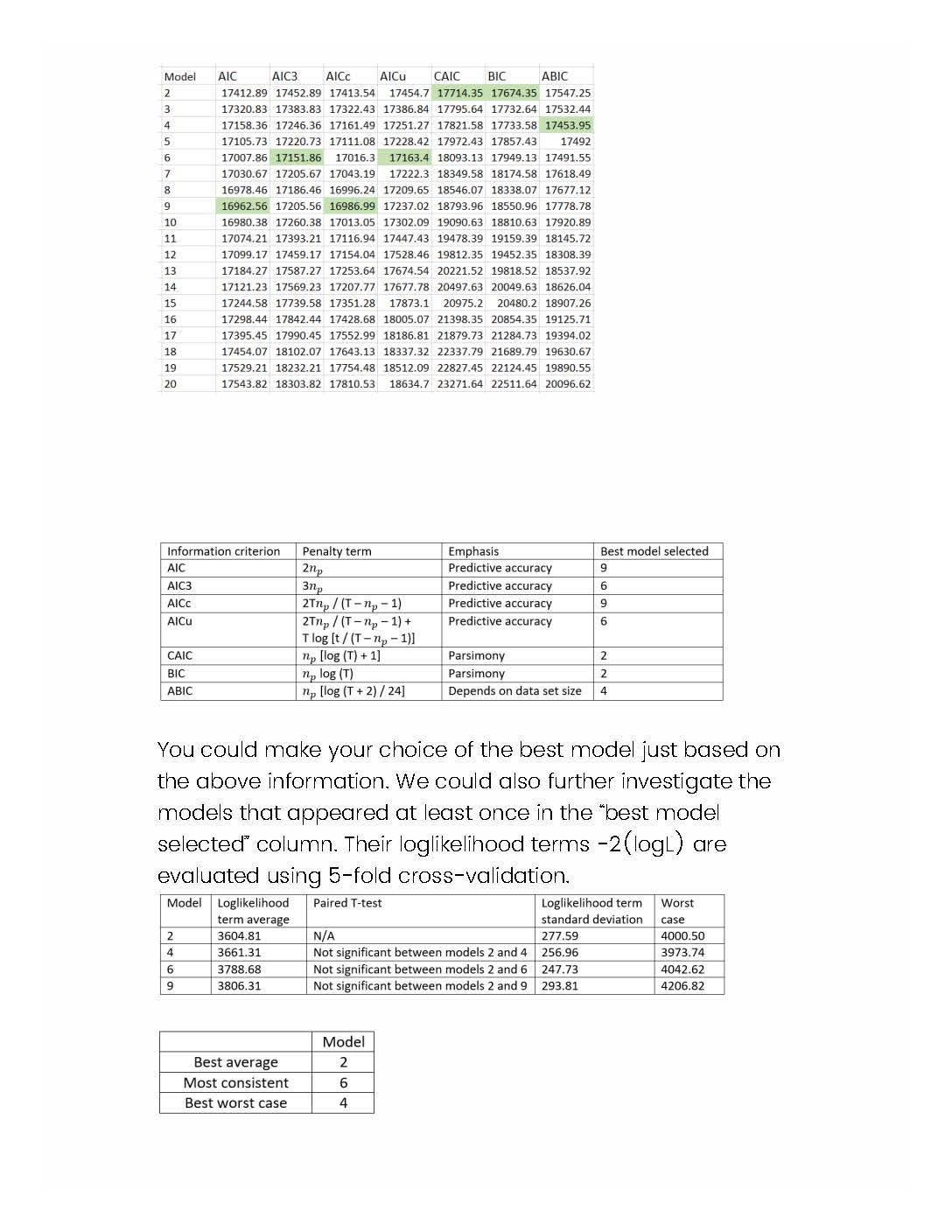}
\end{figure}
\begin{figure}
    \includegraphics[scale=0.75]{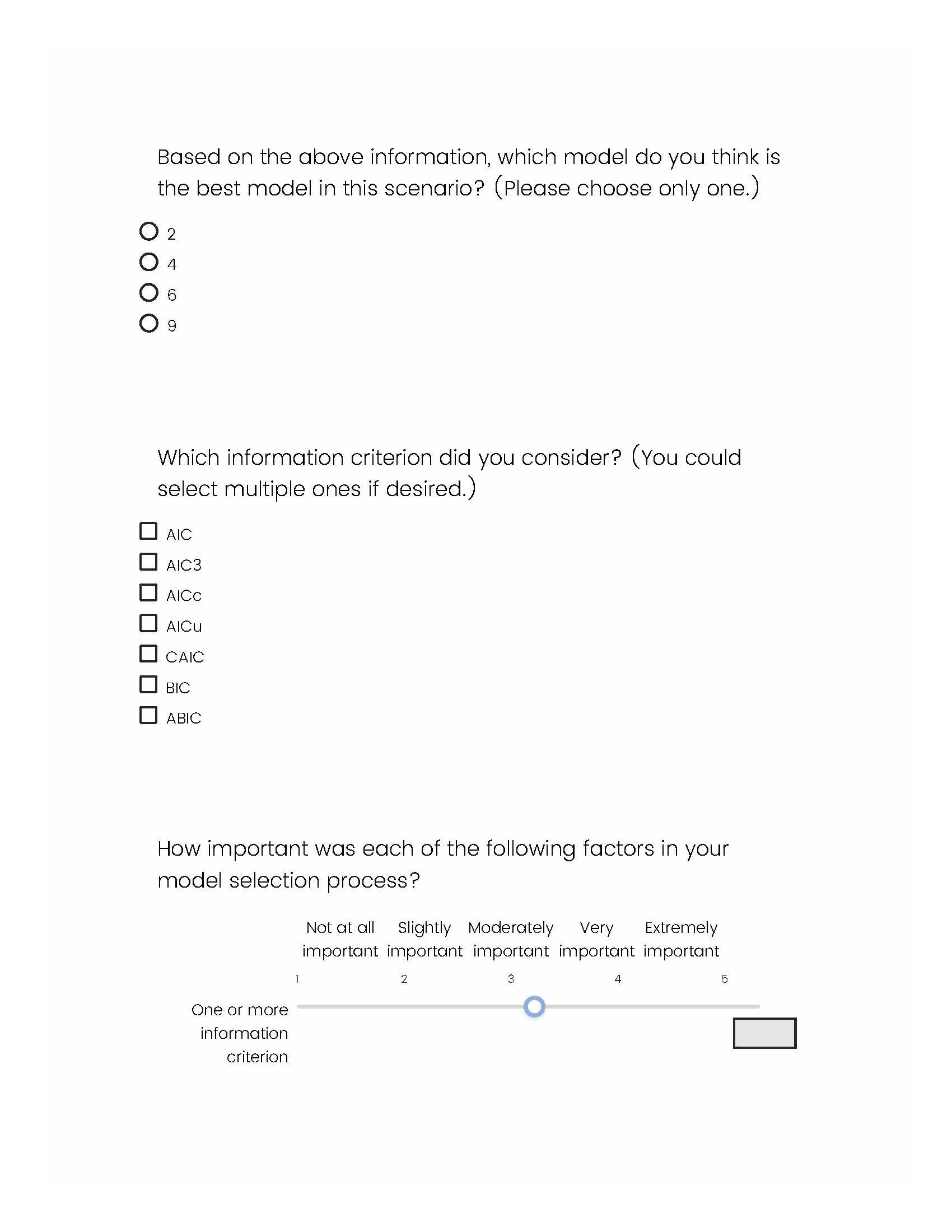}
\end{figure}
\begin{figure}
    \includegraphics[scale=0.75]{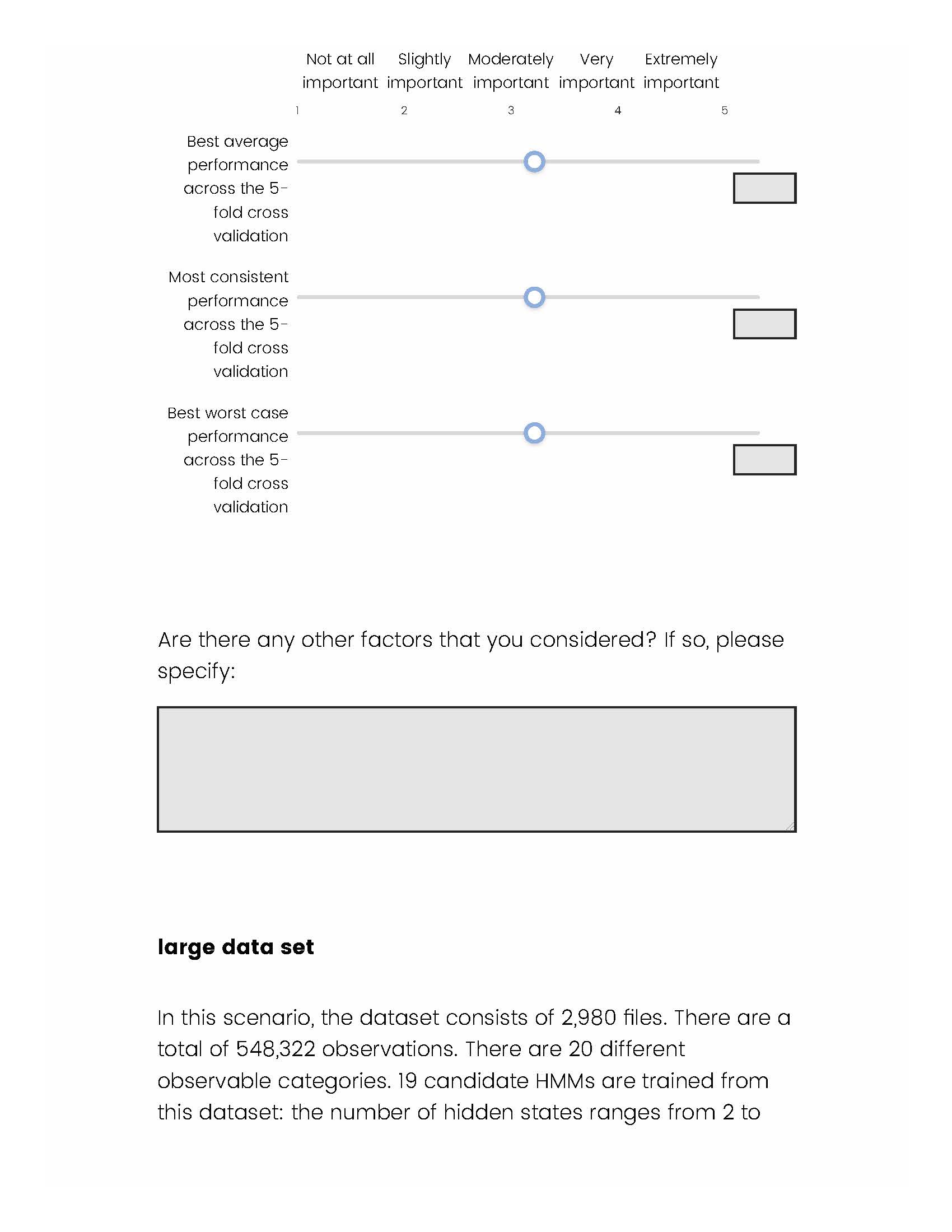}
\end{figure}
\begin{figure}
    \includegraphics[scale=0.75]{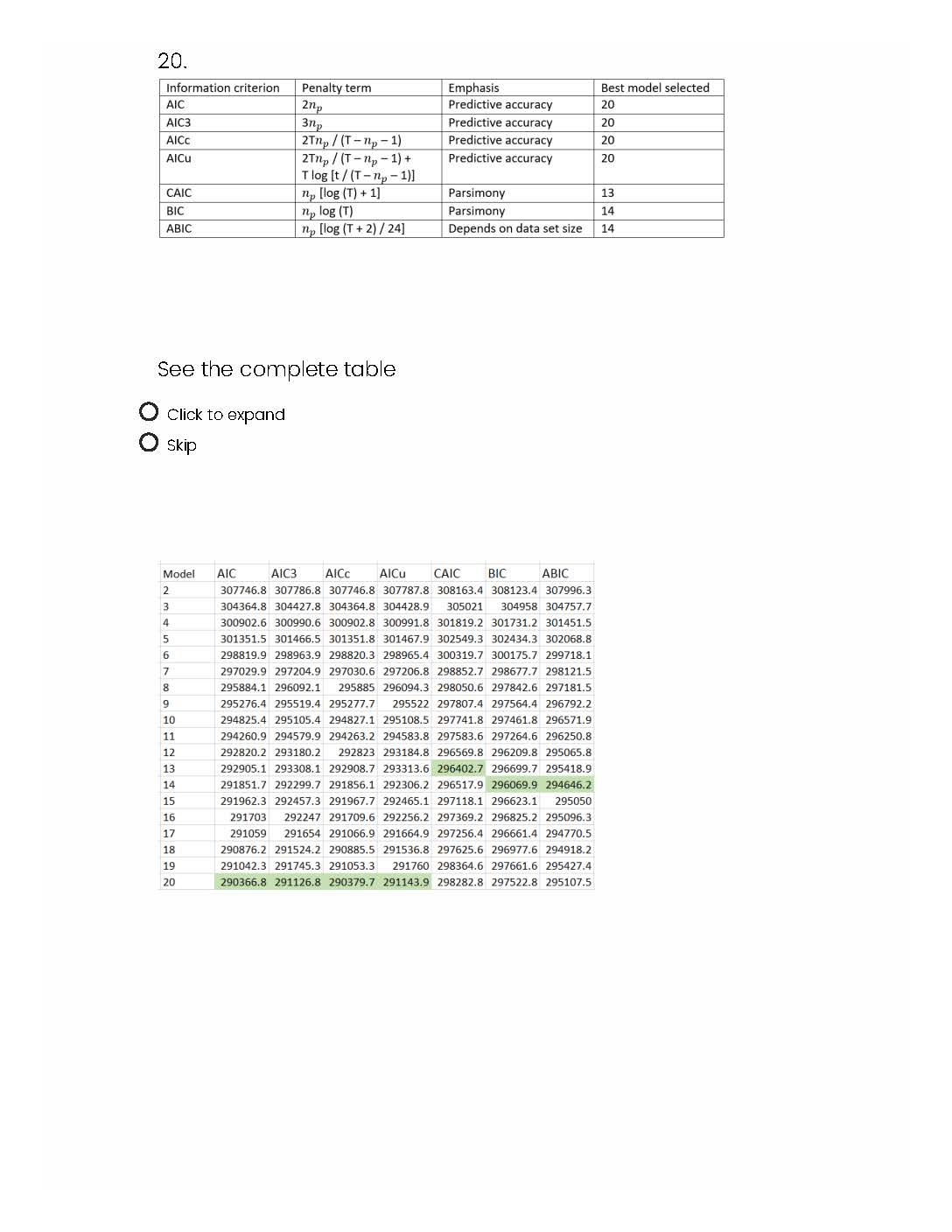}
\end{figure}
\begin{figure}
    \includegraphics[scale=0.75]{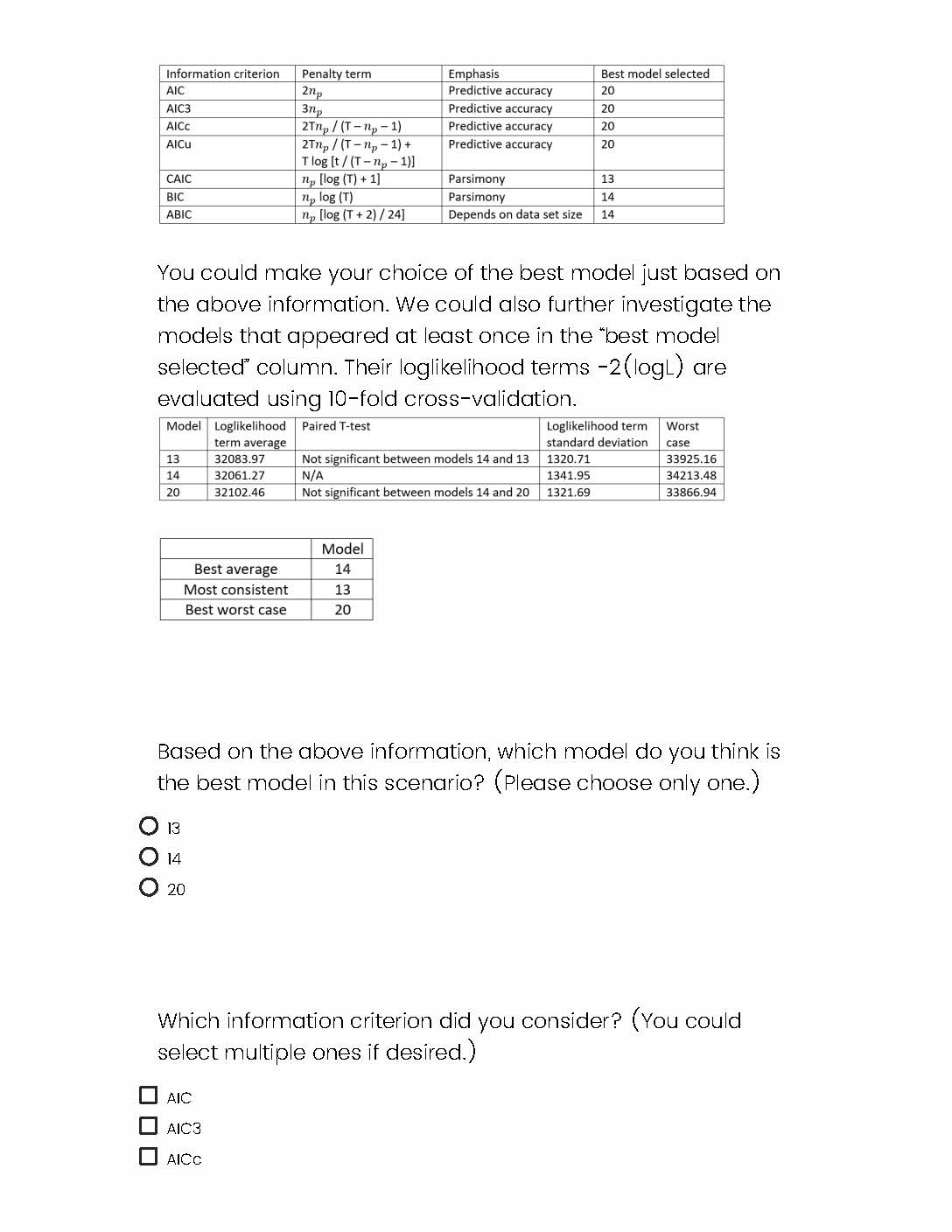}
\end{figure}
\begin{figure}
    \includegraphics[scale=0.75]{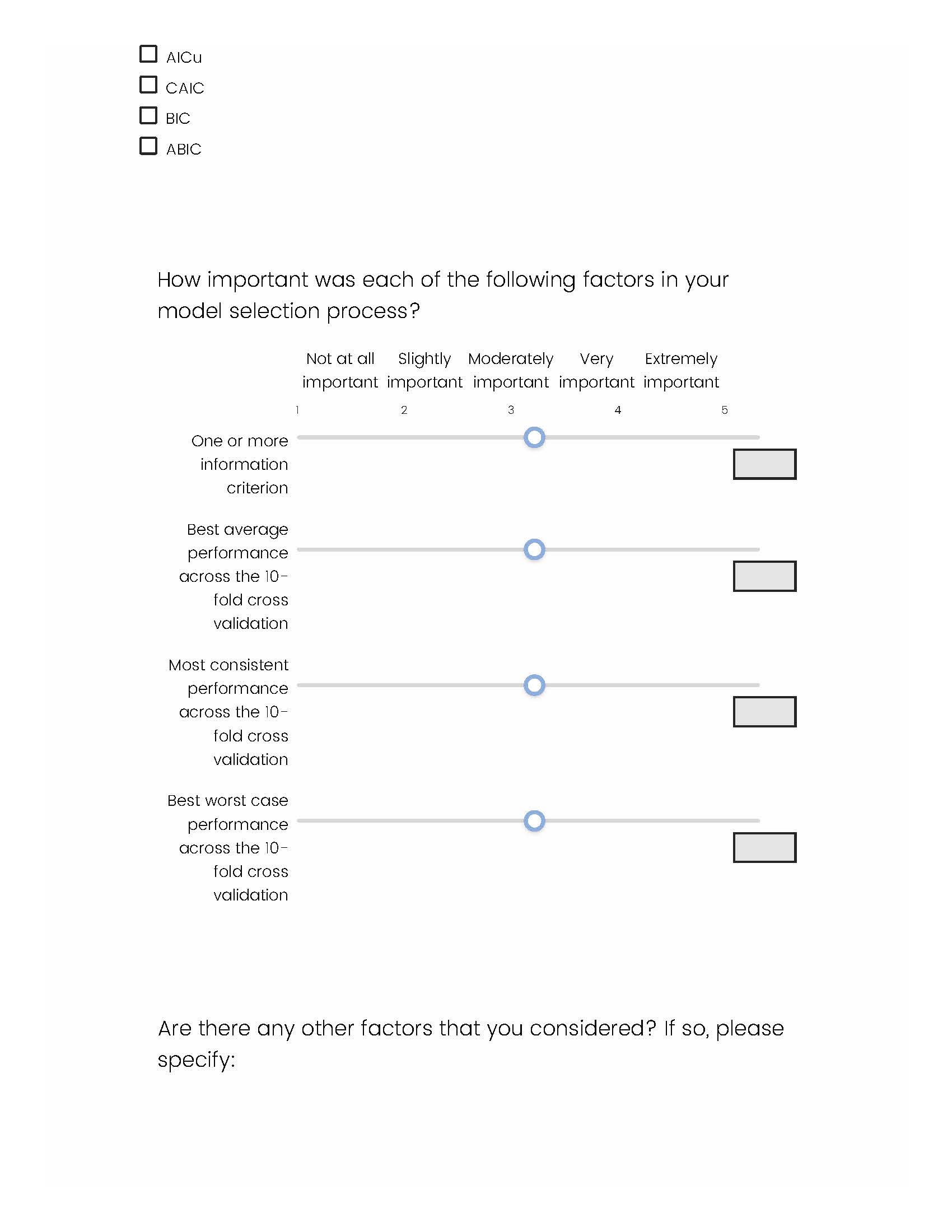}
\end{figure}
\begin{figure}
    \includegraphics[scale=0.75]{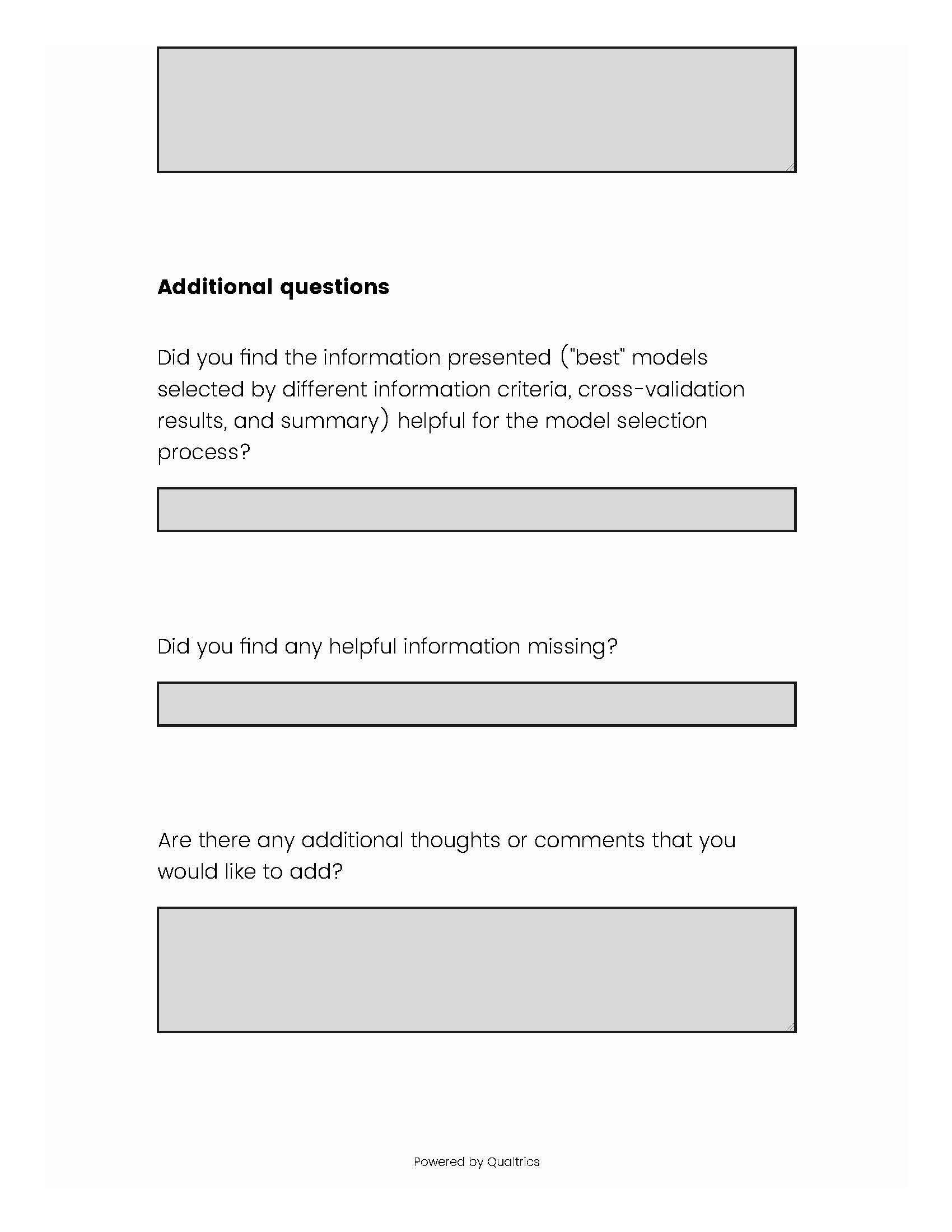}
\end{figure}

\pagebreak
\section{Demographics data}
\includegraphics[scale=0.75]{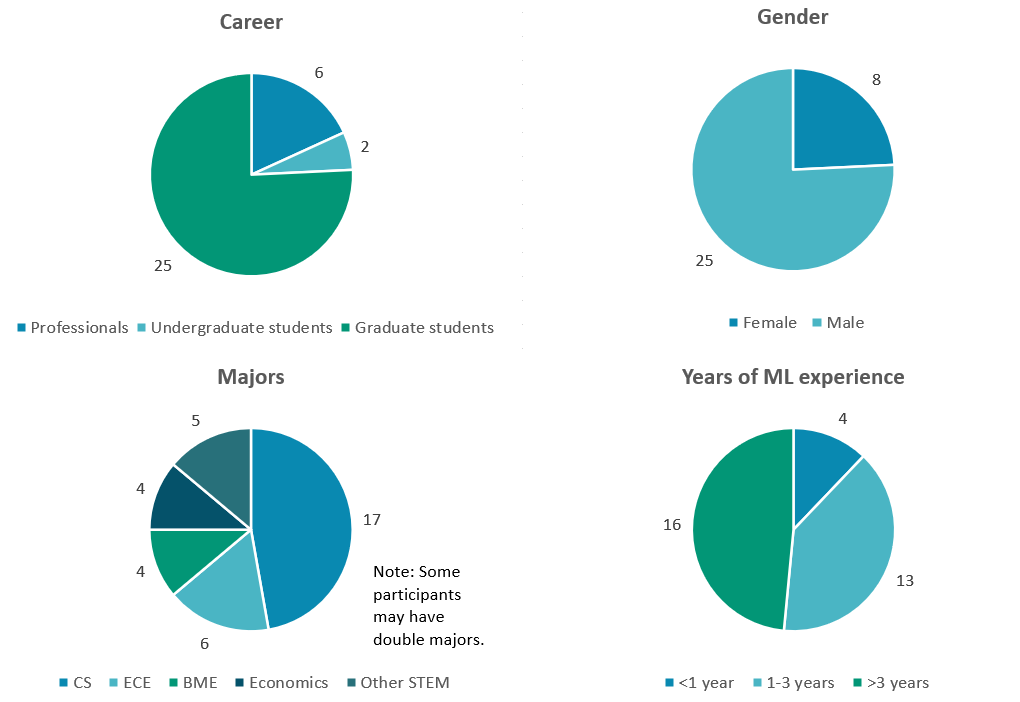}

\section{Correlations}
\begin{figure}[!htb]
    \centering
    \includegraphics[scale = 0.59]{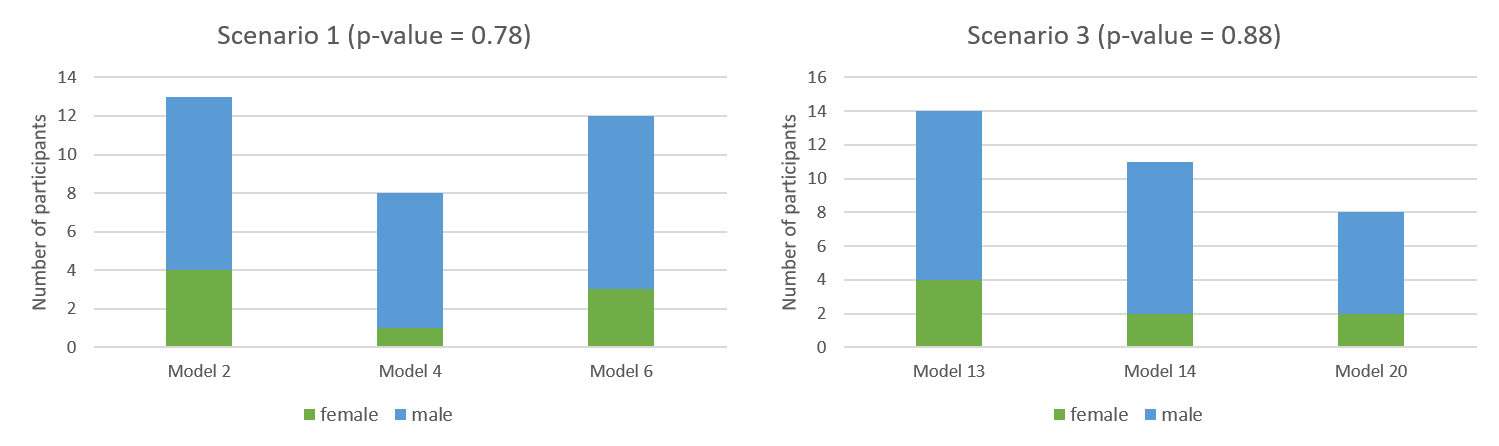}
    \caption{Gender}
    \label{fig:gender}
\end{figure}
\begin{figure}[!htb]
    \centering
    \includegraphics[scale = 0.6]{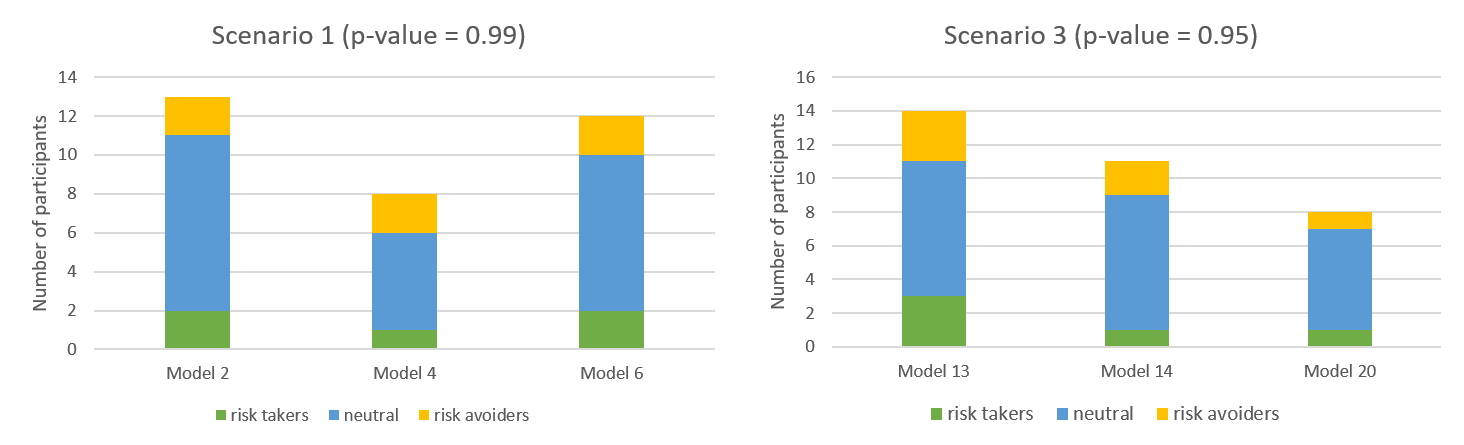}
    \caption{Risk attitude}
    \label{fig:risk}
    \vspace{0.5cm}
    \includegraphics[scale = 0.59]{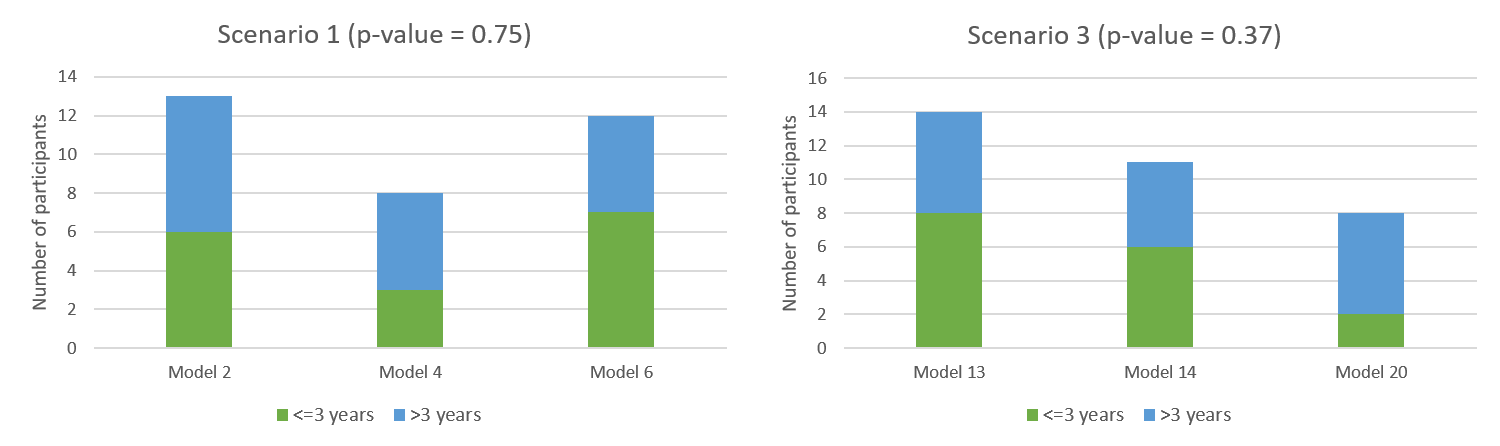}
    \caption{ML experience}
    \label{fig:ml experience}
    \vspace{0.5cm}
    \includegraphics[scale = 0.59]{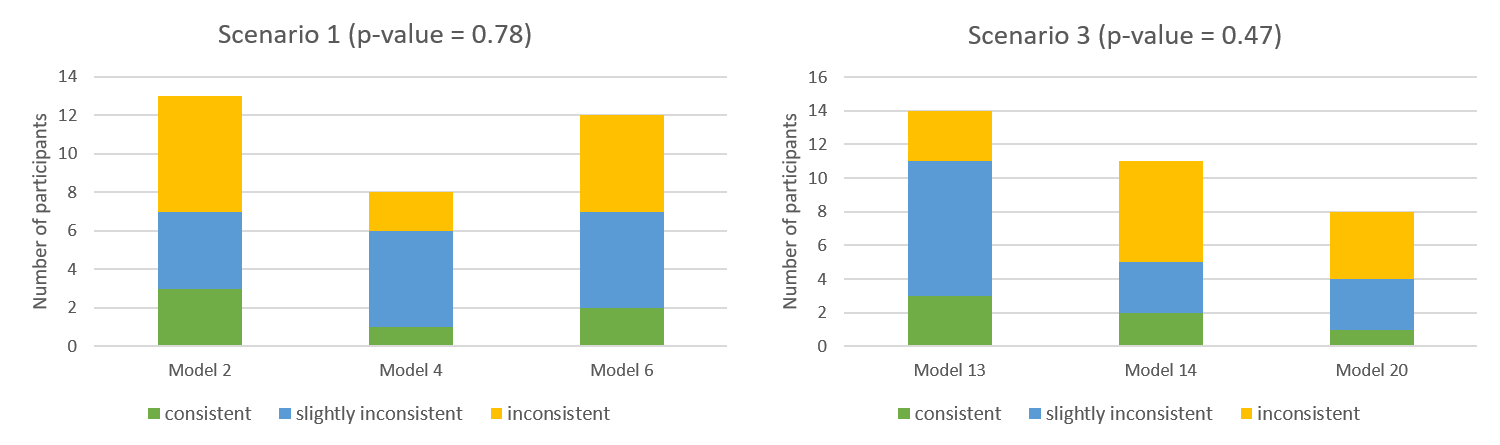}
    \caption{Individual consistency}
    \label{fig:consistency}
\end{figure}
\pagebreak

\section{Example prompts for Large Language Models}
\subsection{Prompt 1}
Now suppose you're a machine learning researcher. You have a data set of 30 files, and you trained 19 Hidden Markov Models based on the files. Their number of hidden states range from 2 to 20. You used multiple information criterion to do model selection. The best model selected by each IC are:

AIC: model with 9 hidden states

AIC3: model with 6 hidden states

AICc: model with 9 hidden states

AICu: model with 6 hidden states

CAIC: model with 2 hidden states

BIC: model with 2 hidden states

ABIC: model with 4 hidden states

You also ran cross-validation tests on model 2, 4, 6, and 9. The “best average” metric selects the model with the lowest negative loglikelihood term across the folds. The “most consistent” metric selects the model with the smallest standard deviation. Additionally, for each model, the largest negative loglikelihood term value among the folds is considered the “worst case.” The "best worst case" metric selects the model with the smallest worst-case value. In this case, the results are:

Best average: model with 2 hidden states

Most consistent: model with 6 hidden states

Best worst case:  model with 4 hidden states

Given these information, which model would you select as the best model? Why?

\subsection{Prompt 2}
In this case, how important was each of the following factors in your model selection decision?

a. one or more information criterion

b. best average

c. most consistent

d. best worst case

Please rate each factor on a scale of 1 to 5 with 1 being not at all important and 5 being extremely important.

\end{document}